\documentclass[accepted]{uai2025}

\usepackage[american]{babel}
\usepackage{natbib}
\usepackage{mathtools}
\usepackage{booktabs}
\usepackage{tikz}
\usepackage{subcaption}
\usepackage{pgfplots}
\usepackage{url}
\usepackage{nicefrac}
\usepackage{xcolor}
\usepackage{amsmath}
\usepackage{amsfonts} 
\usepackage{amssymb}
\usepackage{amsthm}
\usepackage{dsfont}
\usepackage{algorithmic}
\usepackage{algorithm}
\pgfplotsset{compat=1.18}

\newcommand{\bX}{\mathbf{X}}
\newcommand{\bx}{\mathbf{x}}

\newcommand{\argmax}[1]{\underset{#1}{\operatorname{arg}\!\operatorname{max}}\;}
\newcommand{\argmin}[1]{\underset{#1}{\operatorname{arg}\!\operatorname{min}}\;}
\newcommand{\one}[1]{\mathds{1}_{\left\{#1\right\}}}

\newtheorem{proo}{Proof}
\theoremstyle{plain}
\newtheorem{theorem}{Theorem}[section]
\newtheorem{proposition}[theorem]{Proposition}

\theoremstyle{definition}
\newtheorem{definition}[theorem]{Definition}

\newtheorem{remark}[theorem]{Remark}

\title{EERO: Early Exit with Reject Option for Efficient Classification with limited budget}

\author[1,2]{\href{mailto:<florian.valade@univ-eiffel.fr>?Subject=Your UAI 2025 paper}{Florian Valade}}
\author[1]{Mohamed Hebiri}
\author[3]{Paul Gay}

\affil[1]{LAMA \\
Université Gustave Eiffel}
\affil[2]{Fujitsu}
\affil[3]{Université De Pau Et Des Pays De L'adour}

\begin{document}
\maketitle


\begin{abstract}
The increasing complexity of advanced machine learning models requires innovative approaches to manage computational resources effectively. One such method is the Early Exit strategy, which allows for adaptive computation by providing a mechanism to shorten the processing path for simpler data instances. 
In this paper, we propose EERO, a new methodology to translate the problem of early exiting to a problem of using multiple classifiers with reject option in order to better select the exiting head for each instance. We calibrate the probabilities of exiting at the different heads using aggregation with exponential weights 
to guarantee 
a fixed budget. We consider factors such as Bayesian risk, budget constraints, and head-specific budget consumption. 
Experimental results demonstrate that our method achieves competitive compromise between budget allocation and accuracy.
\end{abstract}

\section{Introduction}
\label{sec:intro}

\begin{figure*}[t]
    \resizebox{1\textwidth}{!}{\includegraphics[width=0.9\textwidth, angle=0]{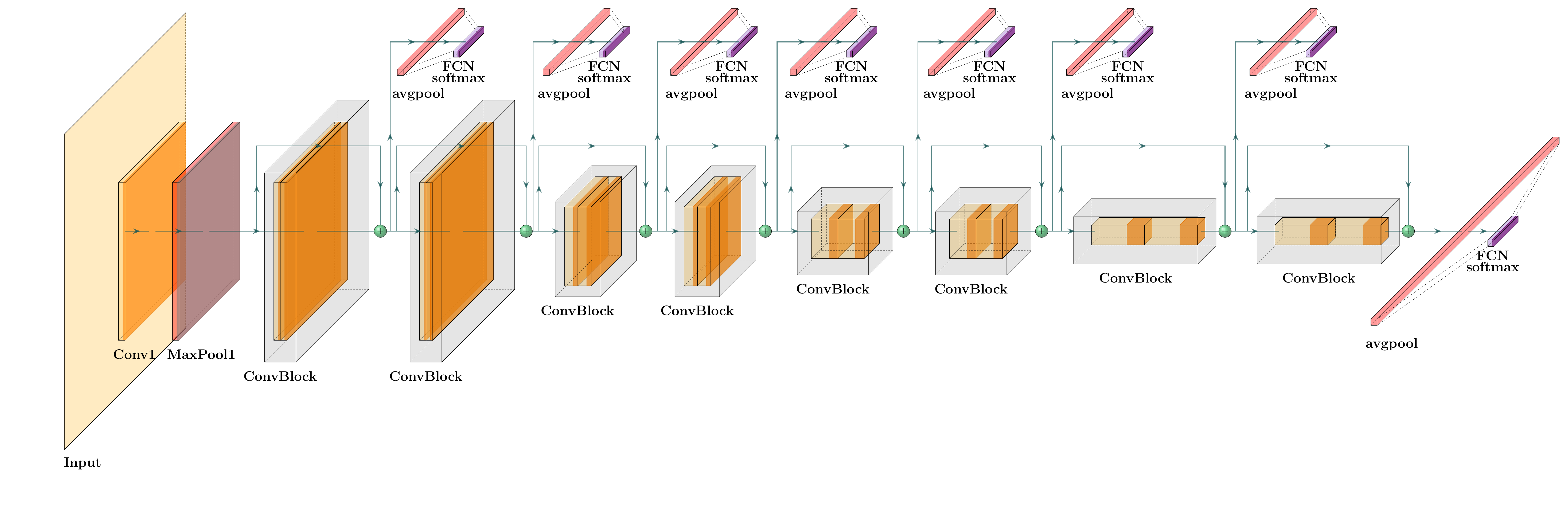}}
    \centering
    \caption{Illustration of the Early Exit principle in a convolutional architecture.}
    \label{fig:resnet}
\end{figure*}

Nowadays, vision models are increasing in size rising the issue of their complexity and computation costs. There exist different strategies to train lighter deep learning networks, such as quantization and pruning~\citep{liang2021pruning}, distillation~\citep{touvron2021training}, and dynamic inference where the network adapts its topology on the fly to the input data~\citep{han2021dynamic}. 
Among them, Early Exit~\citep{laskaridis2021adaptive} is an orthogonal approach which aims at adapting the amount of computation to each input data point, exploiting that most neural networks can be approximated as a stack of layers which process the data sequentially. The idea is to add auxiliary heads at regular intervals along the network (see Figure~\ref{fig:resnet}) which are able to produce a prediction with the current state of the features. The intuition is that easy cases can be processed with the first few layers. 
Those heads can also be used to analyze model's layers~\citep{chen2020learning,kaya2019shallow} and improve training~\citep{teerapittayanon2016branchynet}.

Despite its advantages, one of the main challenges in applying Early Exit during inference is to determine the appropriate moment to exit for a given input. This objective is intricately connected to assessing confidence in the predictions of neural networks. Importantly it requires to build a rule -- that usually relies on a test that asks whether the score for the prediction is higher or not than a given threshold -- at the level of each auxiliary head of the network and 
one possible ultimate goal may be to calibrate all thresholds appropriately in order to meet an overall objective. Specifically, in our contribution, we focus on \textit{Budgeted Batch classification}~\citep{huang2017multi}, a strategy where a fixed computational budget is allocated for processing a batch of data. The strategy involves calculating thresholds to allocate these resources efficiently across different data points enhancing overall accuracy. In principle, this approach offers improved performance because it allows for the conservation of computational resources on simpler cases, which can then be reallocated to improve accuracy on more complex cases. However, practical investigations into this framework have been limited, with most studies concentrating on architectural design or the reliability of confidence scores. Those few that have addressed Budgeted Batch Classification often make additional assumptions for practicality~\citep{huang2017multi} or utilize less-than-ideal algorithms without fully exploring the associated mathematical challenges~\citep{wang2021not}.

\paragraph{Contributions}
In this work, we present EERO (Early Exit with Reject Option), a novel framework designed to augment Early Exit strategies in Budgeted Batch Classification for inference tasks. The EERO methodology progresses through a series of steps: Initially, we train auxiliary heads to refine decision-making. Subsequently, we translate the budget limitation to probabilities of classification at each head based on aggregation with exponential weights~\citep{Dalalyan_2008}. We then use a calibration set and those probabilities to calibrate head specific exit thresholds. We based this calibration of the thresholds on learning with rejection option arguments~\citep{Chow57}. (Also named \emph{learning with abstention} and later \emph{selective learning} in the literature.)
Finally, for each data point and at the level of each head we compute a score. If this score exceeds the threshold, we take this head output as the final prediction. Otherwise, we proceed with the following head in a similar way. 

Our key contributions are: (i) Developing an optimal classification process within a specified GFlops\footnote{Flops: floating point operations per second.} budget, ensuring efficient data processing while balancing accuracy; (ii) Adapting and generalizing EERO to various architectures like ResNet~\citep{he2016deep}, ConvNext~\citep{liu2022convnet}, and MSDNet~\citep{huang2017multi}, suitable for multiple auxiliary heads; (iii) Validating our approach through extensive benchmarks on CIFAR-100 and ImageNet datasets, proving its effectiveness in maintaining resource constraints, reducing computational load, and enhancing model accuracy.

The paper is organized as follows: Section.~\ref{sec:relatedWork} reviews relevant literature. Section.~\ref{sec:Methodo} introduces our proposed EERO method, including its statistical framework (Section~\ref{subsec:statisticalFramework}) and detailed methodology (Section~\ref{subsec:rejectoptionlearning}). Section.~\ref{sec:experiments} presents experimental results, applying EERO on ImagrNet.

\section{Related work}
\label{sec:relatedWork}
One of the first works on Early Exit \citep{teerapittayanon2016branchynet} proposed to use auxiliary heads and a weighted sum of losses. The focus was essentially on the design and the positions of the auxiliary heads. MSDNet~\citep{huang2017multi} is an architecture where the heads are carefully designed so that early features which are assumed to be unsuitable for classification are refined. Although this leads to better performances, the complex resulting heads reduce their number, and therefore the flexibility of the model, and makes it more difficult to adapt to new architectures. Overall, the position and the number of the auxiliary heads remains an open discussion for convolutional networks~\citep{lin2022closer} and transformers~\citep{bakhtiarnia2021multi}.

Determining the optimal timing for an early exit in neural network inference is a crucial aspect of efficient model design. A widely used method is the \textit{Threshold-Based Approach}, where the exit decision is based on a predefined threshold on auxiliary heads output probabilities, focusing on different metrics of the probability distribution~\citep{huang2017multi, bolukbasi2017adaptive}. Another strategy is the \textit{Patience-Based Strategy}, requiring consecutive heads to agree on a prediction before exiting, thus leveraging sequential predictions for reliability~\citep{zhou2020bert}. Additionally, more sophisticated methods involve \textit{Halting Scores}, where scores are accumulated from each layer's output, integrating model confidence into the exit process~\citep{figurnov2017spatially}. In complex scenarios, \textit{Reinforcement Learning Techniques} are employed to finely balance computational efficiency and prediction accuracy, especially in dynamic inference settings~\citep{wang2018skipnet, wu2018blockdrop}.

In the context of \textit{Budgeted batch classification}, thresholds are not determined for individual images but for a batch as a whole, facilitating a more strategic deployment of computational resources. To date, the literature presents two notable algorithms addressing this problem in the realm of Early Exit strategies. The authors in~\citep{wang2021not} employ a genetic algorithm to optimize these thresholds.
Conversely, the work~\citep{huang2017multi} operates under the assumption that each exit point in the network has an equal and predetermined likelihood of accurately classifying an image. While this simplifies the optimization problem, it introduces an additional hyper-parameter that is not inherently related to computational complexity.

In addition, other contexts have been well studied in the literature, such as model cascading or learning-to-defer. Model cascading involves using separate, independent models of increasing complexity, where each model decides whether to classify an instance or defer to the next model in the sequence~\citep{jitkrittum2024doesconfidencebasedcascadedeferral, gupta2024languagemodelcascadestokenlevel}. Learning to defer, on the other hand, transfers decision-making to human operators when the model's confidence is low~\citep{okati2021differentiablelearningtriage, charusaie2024unifying}. Our work differs by presenting a comprehensive formulation that is grounded on the principles of reject option learning theory on a multi headed classification model with a focus on minimizing cost.

Learning with the reject option, a method to abstain from predictions under uncertainty, is crucial in our work. Initially explored in~\citep{Chow57} and advanced through conformal prediction~\citep{Vovk99IntroCP,Vovk_Gammerman_Shafer05}, this concept has evolved significantly~\citep{Herbei_Wegkamp06,Naadeem_Zucker_Hanczar10,Grandvalet_Rakotomamonjy_Keshet_Canu09,Yuan_Wegkamp10,Lei14,cortes2016learning,Denis_Hebiri19}. It generally encompasses strategies for predefined coverage, rejection rates, or a balance of both. In EERO, rejection is adapted at the head level, deciding if an instance $\bx$ should exit early or continue processing. This application to deep learning for energy efficiency is novel, though the reject option has been previously applied to various learning problems~\citep{denis2020regression,DenisHebiriNjikeSibertActive22}.

\section{Method}
\label{sec:Methodo}
In this section, we introduce the Early Exit with Reject Option (EERO) framework, applying reject option learning theory to enable efficient early exits in neural networks.
We detail the statistical underpinnings and describe how EERO strategically manages resources to balance accuracy with computational expenditure. This approach not only optimizes performance but also ensures strict compliance with predefined computational constraints.

\subsection{Statistical framework -- classification with reject option}
\label{subsec:statisticalFramework}
This section describes the mathematical framework for Early Exit. Let $(\bX, Y )$ be a random couple distributed according to a distribution $\mathbb{P}$ on $\mathcal{X}\times[K]$, where $[K]:= \{1, \dots, \textit{K}\}$.

Here, $\mathcal{X} \subset \mathbb{R}^d $ is the feature space, and $Y$ is the label corresponding to the feature $\mathbf{X}$. We focus on the problem of $K$-class classification with $K \geq 2$ and one of our goals is to build a prediction rule $g :\mathcal{X} \to [K] $ that reduces the misclassification risk $ \mathbb{P}\left( g(\bX) \neq Y \right)$. This risk is minimized by the Bayes rule $g^*$ that is given, in the multi-class setting, for all $\mathbf{x}\in \mathcal{X}$ by
\begin{equation}
    \label{eq:BayesRule}
g^*(\bx) = \argmax{k=1,\ldots,K } p_k(\bx) \enspace ,
\end{equation}
where $ p_k(\bx) = \mathbb{P}\left( Y = k | \bX=\bx \right) $ are the conditional probabilities. Because the distribution $\mathbb{P}$ is unknown, the Bayes rule itself is unknown, and we need to approximate it. In general, this approximation seeks to maximize the classification accuracy. In our case, the goal is also to reduce the energy consumption of the model, and thus, we assume a constraint on the computation budget available. Notably, we will show that the latter is strongly connected to classification with reject option (classification with abstention) that we describe in Section~\ref{subsec:rejectoptionlearning}.

The framework of classification with reject option assumes that classifiers are allowed to abstain from classifying (on the empirical side, this means that the classifier abstains on a given proportion of the data). Let $\varepsilon \in (0,1)$ be a parameter that denotes the probability of classifying an instance.
The optimal rule that abstains with probability $1- \varepsilon$ is given by:
\begin{definition}
\label{def:optimalrejectrule}
    Let $\varepsilon \in (0,1)$. The optimal classifier with $1-\varepsilon$ rejection rate is defined as
    \begin{multline}
        h_{\varepsilon}^* \in  \argmin{h} \{ \mathbb{P}( \{ h(\bX) \neq Y \} \cap \ \{ h(\bX) \neq \mathfrak{R}\})  \\
            \qquad \text{s.t. } \mathbb{P}(h(\bX) = \mathfrak{R} ) = 1- \varepsilon \} \enspace ,
    \end{multline}
    where $h$ is a classifier that is allowed to reject, that is,
    $h:\mathcal{X} \to [K]\cup\{ \mathfrak{R} \} $ and $\mathfrak{R} $ is the output when the classifier rejects all elements from $[K]$. 
\end{definition}

The opportunity of using the reject option is important in applications where ambiguity occurs between classes, which is often the case when the total amount of classes $K$ is large. There are several ways to handle this reject option. In this paper, we constrain the rejection rate as it is in accordance with the resource limitation (see Sections~\ref{subsec:rejectformalism} and~\ref{subsec:rejectionratesmultihead}).

Now, we aim at providing the explicit expression of the optimal rule given by Definition~\ref{def:optimalrejectrule}. 
Let us denote by $s$, the score function defined for each $\bx\in \mathcal{X}$ by
\begin{equation}
    \label{eq:scoreMax}
    s(\bx) = \max_{k\in [K]} \left\{p_1(\bx), \ldots , p_K(\bx) \right\} \enspace.
\end{equation}
From the definition of the Bayes rule~\eqref{eq:BayesRule}, we can establish the following characterization of the optimal rule.

\begin{proposition}
\label{prop:BayesRejectRule}
Assume the cumulative distribution function (CDF) $F_{s}$ of $s(\bX)$ is continuous. Then, for all $\bx \in \mathcal{X}$
\begin{equation*}
h_{\varepsilon}^*(\bx) = \left\{
    \begin{array}{ll}
        g^*(\bx)
         & \quad\text{if} \quad F_s(s(\bx))  \geq 1-\varepsilon \\
        \mathfrak{R} & \quad \text{otherwise.}
    \end{array} 
    \right.
\end{equation*}
In particular, we have $\mathbb{P}(h_{\varepsilon}^*(\bX) = \mathfrak{R} ) = 1-\varepsilon$.   
\end{proposition}

Proposition~\ref{prop:BayesRejectRule} extends the result established in~\citep{Denis_Hebiri19} (\emph{c.f.} Proposition~1) to multi-class setting. Its proofs can be found in the Sec.~\ref{App:proofs} of the Appendix. The main assumption used to build this result is the continuity assumption on the CDF of $s(\bX)$ and requires that the random variable has no atoms. It ensures that the classifier $h_{\varepsilon}^*$ has a rejection rate exactly $1-\varepsilon$. 
In the next section, we will see that from an empirical point of view, this condition can always be satisfied through randomization.

\subsection{Data-driven procedure}
\label{subsec:rejectoptionlearning}
Before considering the reject option arguments and since we deal with an Early Exit strategy, we train a neural network with $M$ possible exits that correspond to $M-1$ auxiliary heads and the classical output of the last layer, as illustrated in Figure~\ref{fig:resnet}.
To this end, we collect a \emph{labeled} dataset $\mathcal{D}_{n_1}$, that consists of $n_1 \in \mathbb{N}$ \emph{i.i.d.} copies of $(\bX, Y )$, and train for each head $\ell$ an estimator $(\hat{p}_1^{\ell},\ldots,\hat{p}_K^{\ell})$ of the conditional probability vector $(p_1,\ldots,p_K)$.
Since larger $\ell$ means that we go deeper in the network, it is reasonable to assume that for all layers indices $\ell , \ell' \in [M-1]$ with $\ell < \ell'$, the auxiliary head $\ell'$ consumes more resources than the auxiliary head $\ell$ and in general, yet this increase in consumption might come with a better accuracy.

\subsubsection{Classification rule based on reject option}
\label{subsec:rejectformalism}

Our methodology highlights that early exiting can be efficiently performed, borrowing tools from learning with reject option. Let us then define, for each auxiliary head $\ell$, a rejection rate $1- \varepsilon^{\ell}\in (0,1)$ 
-- that will be specified later. In order to specify the prediction rule for each head, we will mimic the optimal rule provided by Proposition~\eqref{prop:BayesRejectRule} using the plug-in principle. This step requires collecting a sample of \emph{unlabeled} instances $\mathcal{D}_N$, that consists in $N$ \emph{i.i.d.} copies of $\bX$.
Since the process is the same at each auxiliary head, let us develop our methodology for a specific head $\ell$.

The goal is to understand whether the head $\ell$ is a good early exit for a given instance $\bx \in \mathcal{X}$. Translating this into
the classification with reject option vocabulary, the question becomes whether classifier 
\begin{equation}
    \label{eq:hatg}
    \hat{g}^{\ell}(\cdot) = \argmax{k\in [K]} \{\hat{p}_k^{\ell} (\cdot) + u_k\} \enspace,
\end{equation}
should classify the instance $\bx$ or reject it. 
The $u_k$ variable is introduced to randomize the $\hat{p}_k^{\ell}$ estimations for a technical reason that we now explain. 
Let us denote by $(u_k)_{k \in[K]}$ \emph{i.i.d.} variables which follow a uniform distribution on $[0,u]$ with $u$ being a non-negative real number which is usually chosen very small. (In practice, we set $u=10^{-5}$.) It ensures that the random variables $p_k^{\ell}(\bX)+u_k$ has no atoms which is the key argument (see proof in appendix~\ref{App:proofs}) 
to control suitably the rejection rate of the produced classifier with reject option.
Formally, this classifier is the empirical counterpart of the classifier with reject option given by Proposition~\ref{prop:BayesRejectRule} and is based on the empirical score function given for all $\bx$ by
\begin{equation}
\label{eq:EmpiricalScoreMax}
    \hat{s}^{\ell}(\bx) = \max_{k\in [K]} \left\{\hat{p}_1^{\ell}(\bx) + u_1 , \ldots , \hat{p}_K^{\ell}(\bx) + u_K\right\}\enspace.
\end{equation}
Then we can derive the plug-in estimator of the classifier with reject option $h_{\varepsilon}^*$ at the level of the $\ell$-th auxiliary head.
\begin{definition} 
\label{def:empirClassifierWithRejectOption}
For all $\bx \in \mathcal{X}$
\begin{equation*}
\hat{h}_{\varepsilon^{\ell}}^{\ell}(\bx) = \left\{
    \begin{array}{ll}
        \hat{g}^{\ell}(\bx)
         & \quad\text{if} \quad \hat{F}_{\hat{s}^{\ell}}(\hat{s}^{\ell}(\bx))  \geq 1-\varepsilon^{\ell} \\
        \mathfrak{R} & \quad \text{otherwise,}
    \end{array} 
    \right.
\end{equation*}
where conditional on the dataset $\mathcal{D}_n$, we denote by $\hat{F}_{\hat{s}^{\ell}}(t) = \frac{1}{N} \sum_{\bX \in \mathcal{D}_N} \one{\hat{s}^{\ell}(\bX) \leq t }$ the empirical CDF of $\hat{s}^{\ell}(\bX)$ on the \emph{unlabeled} sample $\mathcal{D}_N$.
\end{definition}
\begin{remark}
    The score function $\hat{s}^{\ell}$ in the above definition can be replaced by any suitable metrics providing an estimation of the prediction confidence. We tried entropy-based confidence and breaking ties~\citep{luo2005active} (\emph{i.e.,} the difference between the two highest scores) in our experiments and selected the latter which performed slightly better.
\end{remark}
According to the above definition, the classifier $\hat{h}_{\varepsilon^{\ell}}^{\ell}$ abstains when it is not confident about the classification.
In our case, having $\hat{h}_{\varepsilon^{\ell}}^{\ell}$ assigns the output $\mathfrak{R}$ to a given instance $\bx$ means that the observation should not be treated by $\hat{h}_{\varepsilon^{\ell}}^{\ell}$, but rather should be delayed to next auxiliary head treatment. In contrast, when $\hat{h}_{\varepsilon^{\ell}}^{\ell}(\bx) = \hat{g}^{\ell}(\bx)$, then we use the early exit at the head $\ell$ and the observation $\bx$ does not go through the rest of the network, thus reducing the computation. We can establish the following result whose proof can be found in Section~\ref{App:proofs} of the Appendix.

\begin{proposition}
    \label{propo:EmpiricalRejectionRate}
    For all $\ell\in[M]$ and all  $\varepsilon^{\ell} \in ( 0,1)$, there exists a constant $C > 0$ such that, whatever the distribution $\mathbb{P}$ of the data and whatever the estimators $\hat{s}^{\ell}$ of the score function $s$ we consider, we have 
\begin{equation*}
\left| \mathbb{P}\left( \hat{h}_{\varepsilon^{\ell}}^{\ell}(\bX) =  \mathfrak{R} \right) - (1-\varepsilon^{\ell}) \right| \leq \frac{C}{\sqrt{N}} \enspace.
\end{equation*}
\end{proposition}
The above result confirms that the rejection rate of the classifier $\hat{h}_{\varepsilon^{\ell}}^{\ell}$ is indeed of the right order. 
However, this result suggests that the head $\ell$ might reject more data than it should (by a proportion of order $C/\sqrt{N}$ which is not suitable from the budget perspective. In order to solve this issue, it is sufficient to impose a smaller rate of rejection. More precisely, if we replace $\varepsilon^{\ell}$ by $\tilde{\varepsilon}^{\ell} = \varepsilon^{\ell} + C/\sqrt{N}$ when we run our algorithm, we force the rejection rate to be less than $1-\varepsilon^{\ell}$. From our numerical study, we observed that $C= \varepsilon^{\ell}$ leads to good results in order to ensure the budget limitation. 
\begin{remark}
Our methodology, applicable for semi-supervised learning, capitalizes on both labeled and unlabeled data, making it ideal when acquiring labels is costly. If only labeled data is available, we advise splitting the dataset, facilitating budget control theoretically.
\end{remark}

\begin{remark}
\label{rq:UnrespectedBudgetMultiHead} 
    The result in Proposition~\ref{propo:EmpiricalRejectionRate} is valid when $M>2$ only in the situations where the sets of classified instances by all heads are disjoint, meaning that each image is rejected by all heads but one. Since this is unlikely in practice, we need to adjust the rejection rate at the level of the head $\ell$ based on the rejection rates of the earlier heads. We will elaborate on this in Section~\ref{subsec:rejectoptionmulti}.
\end{remark}

\subsubsection{Calibration of the rejection rates}
\label{subsec:rejectionratesmultihead}

To fully exploit the layered complexity of deep learning networks, our EERO methodology must handle multiple exits.
We incorporate an aggregation with exponential weights~\citep{Cesa-Bianchi_Lugosi_2006,Dalalyan_2008} to optimize the decision-making process at various network depths. This adaptation is crucial for leveraging the diverse representational capabilities of neural networks, ensuring computational efficiency, and maintaining high accuracy across multiple exit points.
We recall that our main constraint here is the maximum allowed budget $B$ in GFlops.

Let us then develop the process to build the vector $\hat{\boldsymbol{\varepsilon}} = (\hat{\varepsilon}^1, \ldots , \hat{\varepsilon}^M)$ of discrete probabilities that provides us the rates of classifying at each head (in other words, the $1- \hat{\varepsilon}^{\ell}$ are the rejection rates).
As inputs, we assume 
\begin{itemize}
    \item we have already trained all heads classifiers that are called $\hat{g}^{1}\ldots \hat{g}^{M}$ (based on the first labeled dataset $\mathcal{D}_{n_1}$);
    \item we have already computed for each of the classifiers $\hat{g}^{\ell}$ an evaluation of its risk
    $\hat{R}^{\ell} = 
    \frac{1}{n_2} \sum_{(\bX ,Y) \in \mathcal{D}_{n_2}} \one{\hat{g}^{\ell}(\bX) \neq  Y }$ based on a second labeled dataset $\mathcal{D}_{n_2}$ that consists in \emph{i.i.d.} copies of $(\bX ,Y)$ -- notice that these error rates are already computed during any classic deep learning training;
    \item we have a prior distribution $\boldsymbol{\pi} = (\pi^1, \ldots, \pi^M)$ on the simplex $\Lambda_{M-1}$ defined as $\pi^{\ell} = \frac{\left( \hat{B}^{\ell} \right)^{-1} }{\sum_{j=1}^M \left( \hat{B}^{\ell} \right)^{-1}  }$, where $\hat{B}^{\ell} $ is the budget required by the head classifier $\hat{g}^{\ell}$ to provide an inference for one instance,
    \item we have fixed the overall budget $B$ we are allowed to use and the size $T$ of the batch of new data points we need to predict. 
\end{itemize}

Our proposal is based on aggregation with exponential weights.
The vector $\hat{\boldsymbol{\varepsilon}} = (\hat{\varepsilon}^1, \ldots , \hat{\varepsilon}^M)\in \Lambda_{M-1}$ is solution in $\boldsymbol{\varepsilon}=(\varepsilon^1, \ldots , \varepsilon^M) $ of the following minimization problem:
\begin{equation}
    \label{eq:pbExpWeigh}
    \min_{\boldsymbol{\varepsilon} \in \mathbb{R}^M} \sum_{\ell=1}^M  \varepsilon^{\ell} \hat{R}^{\ell} + \beta \sum_{\ell =1}^M \varepsilon^{\ell} \log \left( \frac{\varepsilon^{\ell}}{\pi^{\ell}} \right),
\end{equation}
\begin{equation}
    \label{eq:pbExpWeighConstraint}
  s.t.\qquad  \varepsilon^{\ell}\geq 0,\qquad \sum_{\ell =1}^M  \varepsilon^{\ell} =1, \qquad \sum_{\ell =1}^M  \varepsilon^{\ell} \hat{B}^{\ell} \leq \bar{B},
\end{equation}
where $\beta \geq 0 $ is a tuning parameter that controls the strength of the Kullback-Leibler divergence between $\boldsymbol{\varepsilon}$ and $\boldsymbol{\pi}$ and $\bar{B} = B/T$ is the average budget we can spend to infer one instance. Notice that the constraint $\sum_{\ell =1}^M  \varepsilon^{\ell} \hat{B}^{\ell} \leq \bar{B}$ reads as $\sum_{\ell =1}^M  (T\varepsilon^{\ell} )\hat{B}^{\ell} \leq {B}$. In particular, $T \varepsilon^{\ell} $ interprets as the number of data points that should be classified by the head $\ell$ so that the total budget remains less (or equal) than the allocated budget $B$.
Considering the Lagrangian of this problem, we can exhibit the following form of the probability vector $\hat{\boldsymbol{\varepsilon}}$.
\begin{proposition}
\label{prop:rejectionratedefinition}
For all $\ell\in [M]$, the $\ell$-th coordinates of the rejection rates vector $\hat{\boldsymbol{\varepsilon}}$ is given by
\begin{equation*}
    \hat{\varepsilon}^{\ell} = \frac{\pi^{\ell} \exp{\left\{ - \frac{\hat{R}^{\ell} + \hat{\mu} \hat{B}^{\ell}}{\beta} \right\} }}{ \sum_{j=1}^M \pi^j \exp{\left\{ - \frac{\hat{R}^j + \hat{\mu} \hat{B}^j}{\beta} \right\}} } \enspace,
\end{equation*}
where $\hat{\mu} = \max \{0, \bar{\mu} \}$, with $\bar{\mu}$ being solution of
\begin{equation*}
    \sum_{\ell=1}^M (\bar{B} - \hat{B}^{\ell}) \pi^{\ell} \exp{\left\{ - \frac{\hat{R}^{\ell} + \bar{\mu} \hat{B}^{\ell}}{\beta} \right\}} = 0 \enspace.
\end{equation*}
\end{proposition}
The only tuning parameter in the above procedure is the temperature $\beta$. Higher values force the probability vector $\hat{\boldsymbol{\varepsilon}}$ to get closer to the prior distribution $\boldsymbol{\pi} $ that is created to take into account the budget required by each head to produce a prediction. Last but not least, The empirical risk factors $\hat{R}^{\ell}$ in Equation~\eqref{eq:pbExpWeighConstraint} allow the rejection rates $\hat{\boldsymbol{\varepsilon}}^{\ell}$ to depend on the head performances on the training data, \emph{i.e.,} layers with good classifiers will be selected more often. 

One main advantage of aggregation with exponential weights is that it is supported by strong theoretical performance.
Let us focus on the misclassification risk $\mathcal{R}(g) := \mathbb{P}\left( g(\bX) \neq Y\right)$. 
Following the analysis in~\cite{Dalalyan_2008,RigolletTsyb12ExponentialWeighting}, we can show the following result.
\begin{theorem}
\label{thm:OracleInequality}
    Consider $\hat{g}^{\text{EERO}}$ the classifier resulting from our aggregation procedure (whose formal definition is given in\eqref{proofdef:EERO}). Conditionally on $\mathcal{D}_{n_1}$, the data used to train the heads $\hat{g}^{1}\ldots \hat{g}^{M}$, we have 
$$
\mathcal{R} ( \hat{g}^{\text{EERO} }) \leq \inf_{\ell \in [M] : \ \hat{B}^{\ell} \leq \bar{B} }  \left\{ \mathcal{R} (\hat{g}^{\ell} ) +   \beta     \log (  1/\pi^{\ell} )  \right\}\enspace.
$$
\end{theorem}
Notice that the above Oracle inequality shows that our aggregation procedure $\hat{g}^{\text{EERO} }$ is at least as good (up to a log factor) as the best head that satisfies the budget constraint. In addition, without \emph{prior} knowledge on the budget, one may consider the uniform probabilities $\pi^{\ell} = 1/M $ and recover the classical bound of order $\log(M)$.

\begin{remark}
    Our methodology fits well for the problem of Budgeted Batch Classification. However, the notion of batch is not important for our purpose. As expressed in the above algorithm~\eqref{eq:pbExpWeigh}-\eqref{eq:pbExpWeighConstraint} (see also Proposition~\ref{prop:rejectionratedefinition}), the only information that is required to calibrate the probability of rejections vector $\hat{\boldsymbol{\varepsilon}}$ is the average budget for classifying one instance. 
\end{remark}
\begin{remark}
EERO methodology takes advantage of aggregation with exponential weights to compromise between accuracy and budget consumption. However in the particular case of only one auxiliary head, there is no need for aggregation -- the budget is split between the auxiliary head and the last layer. We have developed such particular case in Appendix~\ref{app:TwoHeads} and run experiments in that case as well using the ResNet-18 architecture illustrated in Figure~\ref{fig:resnet} on CIFAR-100.
\end{remark}

\subsubsection{Heads with reject option}
\label{subsec:rejectoptionmulti}

In the case of multiple heads, it is difficult to ensure the validity of Proposition~\ref{propo:EmpiricalRejectionRate} for all heads (\emph{c.f.} Remark~\ref{rq:UnrespectedBudgetMultiHead}). However, we can guarantee a weaker result that is sufficient to ensure the good control on the allocated budget.
Based on the previous section, we built $M$ head classifiers on one hand and a probability vector $\hat{\boldsymbol{\varepsilon}}$ whose $\ell$-th components gives the rates of classification at the $\ell$-th auxiliary head on the other hand.
The probability vector $\hat{\boldsymbol{\varepsilon}}$ takes into account both the accuracy and the resources of each head and allows achieving the suitable control on the budget. However, in the case of multiple auxiliary heads, a careful analysis of our methodology imposes some modifications according to the sequential calibration of the probabilities of classification.

The calibration of the classification rates of all heads is based on the estimation of the CDF $F_s$ of the score function. Since there are $M-1$ auxiliary heads, we estimate $M-1$ times the function $F_s$. Importantly, all these estimates are built on the same dataset $\mathcal{D}_N$.
While the calibration at the level of the first head is perfectly valid, the calibration of the classification rate starting from the second head needs to be adjusted to get the $\hat{\varepsilon}^{\ell}$ classification rate. In particular, for all $\ell \in \left\{ 2, \ldots, M-1 \right\}$, we enforce a higher classification rate as
\begin{equation}
    \label{eq:scoreseq}
\hat{\varepsilon}_{\rm{seq}}^{\ell}  = \sum_{j = 1}^{\ell}\hat{\varepsilon}^{j} \enspace.
\end{equation}
This choice is motivated by the fact that we need, for each head $\ell$, to calibrate the threshold for the rejection rule so that at most a proportion $1-\sum_{j = 1}^{\ell}\hat{\varepsilon}^{j}$ of the data in $\mathcal{D}_N$ is rejected.
If we consider this adjustment together with the correction we have detailed right after Proposition~\ref{propo:EmpiricalRejectionRate}, we can show that for all $\ell\in[M]$, if we replace the probability of classification   $\hat{\varepsilon}^{\ell}$ in Definition~\ref{def:empirClassifierWithRejectOption} by 
\begin{equation}
\label{eq:scoreseqPertub}\tilde{\varepsilon}_{\rm{seq}}^{\ell} = \hat{\varepsilon}_{\rm{seq}}^{\ell} + C/\sqrt{N} \enspace.
\end{equation}
We have that $\mathbb{P}\left(\hat{h}_{\tilde{\varepsilon}_{\rm{seq}}^{\ell}}^{\ell}(\bX) =  \mathfrak{R} \right)     \leq 1- \sum_{j=1}^{\ell} \hat{\varepsilon}^{j} $ so that we can deduce the following result.
\begin{theorem}
\label{thm:controlBudgetheadl}
    For all $\ell \in [L]$, the classification procedure at the head $\ell$ is such that
    \begin{equation*}
    \mathbb{P}\left(\hat{h}_{\tilde{\varepsilon}_{\rm{seq}}^{\ell}}^{\ell}(\bX) =  \mathfrak{R} \right)     \leq  \sum_{j= \ell + 1}^{M} \hat{\varepsilon}^{j} \enspace.
    \end{equation*}
\end{theorem}
The above statement means that after using the head $\ell$, there is at most a proportion $\sum_{j= \ell + 1 }^{M} \hat{\varepsilon}^{j} $ of the data that remains to classify.
This result will be illustrated in the next section through numerical experiments.
\begin{algorithm}
\caption{EERO method: Calibration phase and classification of a batch of data points}
\label{alg:rejectOption}
\begin{algorithmic}[1]
\REQUIRE Batch of data points: $\bX_1,\ldots,\bX_T\in \mathcal{X}$;\\Calibration sets $\mathcal{D}_N$; Model $p$ with $M$ heads, \\Risk of heads: $R$, Allowed budget: $B$
\ENSURE Prediction: $\mathcal{P} \in [K]^T$

\STATE $\boldsymbol{F_s} \leftarrow \text{ComputeCDF}(\mathcal{D}_N)$  \hfill \COMMENT{Eq.~\eqref{eq:EmpiricalScoreMax}}
\STATE $\boldsymbol{\tilde{\varepsilon}_{\rm{seq}}} \leftarrow \text{Aggregation}(M, R, B)$ \hfill \COMMENT{Prop~\ref{prop:rejectionratedefinition}, Eq~\eqref{eq:scoreseq}-\eqref{eq:scoreseqPertub}}
\STATE $\mathcal{P} \leftarrow$ empty list
\FOR{$i = 1$ to $T$}
    \FOR{$\ell = 1$ to $M$}
        \STATE $s^{\ell}, g^{\ell} \leftarrow \text{CompOutput}(\bX_i, p_\ell)$ \hfill \COMMENT {Eq.~\eqref{eq:hatg}-\eqref{eq:EmpiricalScoreMax}}
        \IF{$F_{s^{\ell}}(s^{\ell}(\bx))  \geq 1-\tilde{\varepsilon}_{\rm{seq}}^{\ell}$}
            \STATE $\mathcal{P} \leftarrow \left[ \mathcal{P}, g^{\ell} \right]$
            \STATE break
        \ENDIF
    \ENDFOR
\ENDFOR
\STATE \textbf{return} $\mathcal{P}$
\end{algorithmic}
\end{algorithm}

\section{Experiments}
\label{sec:experiments}
Building upon the methodological foundations established for EERO, this section seeks to empirically substantiate our approach on ImageNet dataset with the ConvNext and MSDNet network architectures. We also compare our EERO approach with MSDNet strategy to compute the exit thresholds. 
Those experiments\footnote{All computations are run on a server with an Intel(R) Xeon(R) Gold 5120 CPU and a Tesla V100 GPU with 32GB of Vram and 64GB of RAM. Code associated with paper can be found in repository here : \url{https://github.com/FlorianVal/Early-Exit-With-Reject-Option}} validate our theory and demonstrate the scalability of EERO across diverse neural network configurations.

\subsection{EERO with multiple heads}
\label{subsec:eeromultihead}
We recall that, after training the heads, the procedure consists of two steps.
The first one focuses on determining the rejection rates at each head. The second one consists in specifying the rejection rule (that is, calibrating the exit thresholds) at the level of each given head, \emph{c.f.,} Definition~\ref{def:empirClassifierWithRejectOption}.

In particular, we exploit the methodology based on aggregation with exponential weights developed in Section~\ref{subsec:rejectionratesmultihead} to specify the rejection rates.

Following the protocol from Section~\ref{subsec:rejectoptionlearning}, EERO is applied to the ConvNext architecture using the ImageNet dataset. The model is adapted with auxiliary heads -- See Appendix~\ref{app:details} for details on extra computational cost induced by training the auxiliary heads. The training is based on pretrained weights and a sum of cross-entropy losses from various exits. For complete hyperparameter details, refer to Appendix Tables~\ref{table:basic-setup} and~\ref{table:training-parameters}.
\begin{figure}
    \includegraphics[width=.95\linewidth]{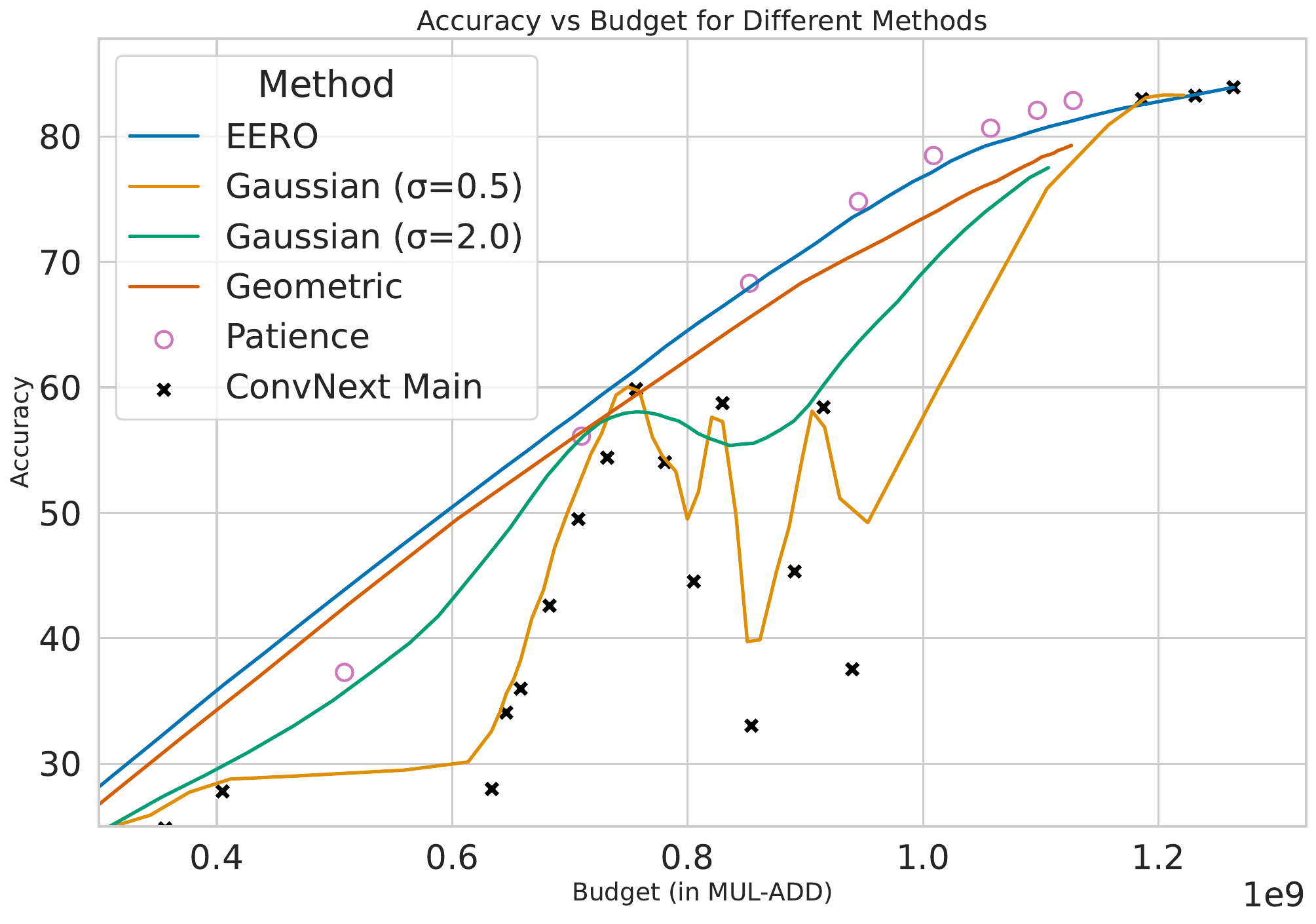}
    \caption{Accuracy \emph{w.r.t.} the budget for our EERO methodology based on Convnext against other known methods such as Patience~\citep{zhou2020bert, zhang-etal-2022-pcee}, Geometric distribution~\citep{huang2017multi, elbayad2020depthadaptivetransformer} or Gaussian~\citep{li2022predictiveexitpredictionfinegrained} distribution of weights on each head. ConvNext Main points correspond to the accuracy of each head alone.}
    \label{fig:convnext_acc}
\end{figure}
\begin{figure}
    \includegraphics[width=.95\linewidth]{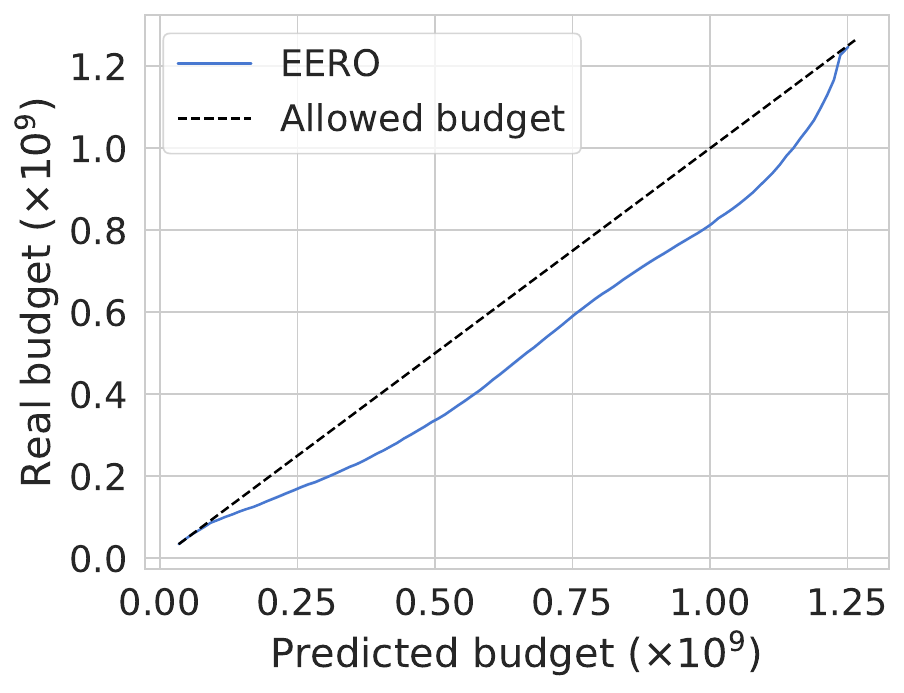}
    \caption{Measured and allowed budget of our algorithm on Convnext. This figure shows that our method accurately follows the budget given by never exceeding it.}
    \label{fig:budget_check}
    
\end{figure}

We plot the results for the different heads and for different computation budgets in Figure~\ref{fig:convnext_acc}.
Several observations can be made. First, we highlight an interesting phenomenon: for a given budget, our proposed multiple heads approach (Blue curve) improves the accuracy from the different exits used independently (black cross). Such observation confirms our statement in Theorem~\ref{thm:OracleInequality}. Moreover, regarding the accuracy of the auxiliary heads alone, placing an exit further in the model does not necessarily result in a better accuracy when the model is not tuned for early exit.

\begin{figure}[b]
    \includegraphics[width=0.95\linewidth]{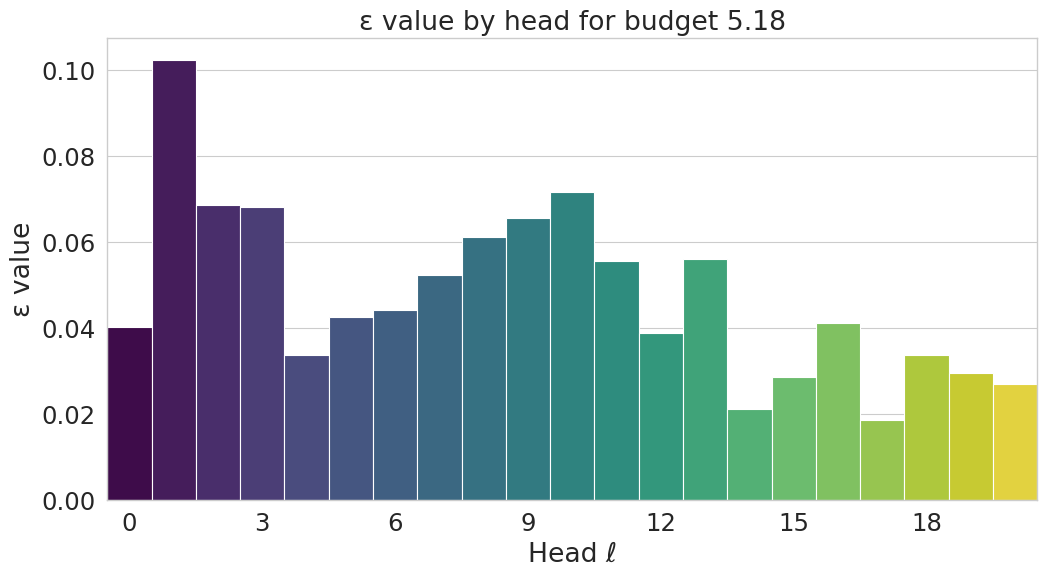}
    \caption{Value of the aggregation weights $\hat{\varepsilon}^{\ell}$ on an intermediate budget for the ConvNext model for EERO. Each bar represents an exit head $\ell$.}
    \label{fig:epsilon_aggreg}
\end{figure}

We further check the budget used by our model in Figure~\ref{fig:budget_check} and show that our methodology respects the allowed budget by always being lower or equal to it, validating our theory from Proposition~\ref{propo:EmpiricalRejectionRate} and Theorem~\ref{thm:controlBudgetheadl}. On some budgets, the gap between the allowed budget and the actual consumption suggests that a better accuracy could be obtained. We also found that the rejection rates $\hat{\varepsilon}^{\ell}$ behave as expected as for example lower budgets favor early heads, and heads with higher risk $\hat{R}^{\ell}$ have lower values, thus decreasing their likelihood to be selected for the final classification score as depicted on Figure~\ref{fig:epsilon_aggreg}. (See also Appendix~\ref{app:numerics}).

\subsection{Comparison with Existing Early Exit Strategies}

We extend our evaluation by comparing EERO with existing early exit strategies, demonstrating its general applicability and effectiveness across different models and scenarios.

\subsubsection{Methods Using specific distributions of Exits}
Some early exit strategies employ predefined distributions, such as Gaussian~\citep{li2022predictiveexitpredictionfinegrained} or geometric~\citep{huang2017multi, elbayad2020depthadaptivetransformer} distributions, to determine the probability of exiting at each head. We show on Figure~\ref{fig:convnext_acc} that these methods lack awareness of the individual risks associated with each head, which can significantly impact overall accuracy. While they may achieve comparable results on networks specifically tuned so that deeper exits yield better accuracy, they are less effective on general architectures. The lack of adaptation to the actual performance of each head can lead to suboptimal allocation of computational resources, potentially favoring less accurate exits. This is highlighted in Figure~\ref{fig:convnext_acc} where our method outperforms specific-distribution methods.

\subsubsection{Patience-Based Strategies}
Patience-based methods \citep{zhou2020bert, zhang-etal-2022-pcee} exit after a certain number of consecutive confident predictions, leveraging the model's consistency across layers. While these strategies can outperform EERO in specific cases, they do not allow for explicit computational budget selection. Budget control is indirect and limited to discrete levels corresponding to different patience values, which restricts fine-grained resource allocation and may not align with strict budget constraints.

\subsubsection{Comparison with MSDNet}

In our experiments (see Appendix ~\ref{fig:MSDNet_acc},~\ref{fig:MDSNet_budget}), we compared EERO with the Multi-Scale Dense Network (MSDNet) \citep{huang2017multi}, a well-known architecture specifically designed for early exiting that uses geometric distribution. The results show that EERO achieves similar accuracy and computational budget consumption as MSDNet (Figure~\ref{fig:MSDNet_acc}). However, there are key differences between the two approaches. MSDNet requires a specialized architecture and tuning to ensure that exits placed deeper in the network provide better accuracy. Moreover, MSDNet does not allow for the direct specification of a computational budget; instead, it requires iterative adjustments of exit probabilities to approximate the desired budget, which can be impractical.
In contrast, EERO is model-agnostic and can be adapted to any network architecture with auxiliary heads. It allows for explicit selection of the computational budget beforehand, ensuring that the resource constraints are strictly met (Figure~\ref{fig:MDSNet_budget}). This flexibility makes EERO more suitable for applications where budget compliance is critical.

\subsubsection{Advantages of EERO}
EERO addresses the limitations of these methods by providing a general framework that is both model-agnostic and budget-aware. By incorporating the risks of each head through our aggregation method (Section~\ref{subsec:rejectionratesmultihead}), EERO optimally allocates computational resources to maximize accuracy under a given budget constraint. This ensures that more computational effort is devoted to heads with better performance, regardless of their position in the network.
Our experimental results (Figure~\ref{fig:convnext_acc}) demonstrate that EERO matches or outperforms the performance of specialized methods without requiring architecture-specific tuning. It provides precise budget adherence (Figure~\ref{fig:budget_check}) and maintains competitive accuracy across various budgets and architectures.

\section{Conclusion}
\label{sec:concl}
In this paper, we presented EERO, a novel mathematical framework devised to construct classification rules for Early Exit in deep learning networks. This framework targets the pivotal challenge of optimizing computational efficiency while enhancing model performance. EERO approaches the problem by modeling it as a classification with a reject option, providing a feasible solution for scenarios involving a single Early Exit. 
For multi-exit scenarios, our innovative approach hinges on an aggregation with exponential weights in order to tune the exiting probabilities. A key strength of EERO lies in its flexibility and generalizability, enabling it to be applied across various model architectures. This adaptability makes EERO a versatile tool, well-suited for diverse applications in the realm of efficient deep learning.

Our experiments on ImageNet and CIFAR-100 using the Resnet-18, MSDNet and the ConvNext architectures respectively demonstrated the efficiency of our framework. The results not only shed light on the intricate workings of multi-head deep neural networks but also offer practical strategies to enhance model accuracy while reducing computational budget. This has important implications for edge computing and making the field of deep learning more sustainable.

\section*{Acknowledgements}

This work was conducted as part of a CIFRE PhD funded by Fujitsu and the French National Association for Research and Technology (ANRT). We would like to thank Fujitsu for their support and collaboration throughout this research.

\bibliography{egbib}

\newpage

\onecolumn
\title{Supplementary Material}
\maketitle
\appendix
\section{Implementation details of experiments}
\label{app:details}

In Tables~\ref{table:basic-setup} and~\ref{table:training-parameters}, we provide the key experimental details for our implementations, including model architectures, datasets, and training parameters. Our implementation is done by freezing the pretrained model and only training the added auxiliary heads. Involving the extra head in our process increases the computational cost by less than $4\%$ and $2.5\%$ in the case of ResNet-18 and ConvNext respectively. In all our experiments we set $\beta = 0.04$ and we mention that our method is not significantly affected by changes of this parameter ; notice that our theory suggests a value of order $1/\sqrt{N}$.

\begin{table*}[h]
\centering
\caption{Basic Setup of Experiments}
\label{table:basic-setup}
\begin{tabular}{|l|l|l|p{5cm}|}
\hline
\textbf{Experiment}        & \textbf{Model}   & \textbf{Dataset} & \textbf{Key Modifications}                                                 \\ \hline
ResNet with CIFAR-100      & ResNet           & CIFAR-100        & Auxiliary heads at 7 positions, Flatten outputs                            \\ \hline
ConvNext with ImageNet     & ConvNext         & ImageNet         & Auxiliary heads at 19 positions, Adaptive Average pooling, Flatten outputs \\ \hline
\end{tabular}
\end{table*}

\begin{table*}[h]
\centering
\caption{Detailed Training Parameters}
\label{table:training-parameters}
\begin{tabular}{|l|p{6cm}|l|}
\hline
\textbf{Experiment}        & \textbf{Training Details}          & \textbf{Calibration Details}                                                                                                                     \\ \hline
ResNet with CIFAR-100      & SGD, LR: 0.003, Batch size: 128 Train data : 49.000 & Calibration set: 1,000 examples \\ \hline
ConvNext with ImageNet     & Pretrained weights, Fine-tuned for 300 epochs, SGD, Momentum: 0.9, Weight decay: \(10^{-4}\), Train data : Full Imagenet & Calibration set: 5,000 examples \\ \hline
\end{tabular}
\end{table*}

\section{Analysis of aggregation weights}
We examine the values of the obtained thresholds $\hat{\boldsymbol{\varepsilon}}$ as it offers interesting insights. 
We first display the values for all the layers and for different budgets in Table~\ref{epsilon_table}. 
\label{app:numerics}
\begin{table*}[ht]
\centering
\begin{tabular}{c|c|c|c|c|c|c|c|c}
\toprule
& & & \multicolumn{2}{c|}{$6.55 *10^{12}$ Flops} & \multicolumn{2}{c}{$3.21 *10^{13}$ Flops} & \multicolumn{2}{|c}{$5.64 *10^{13}$ Flops} \\
\midrule
$\ell$ & $\hat{R}^{\ell}$ & $\pi^{\ell}$ & $\hat{\varepsilon}^{\ell}$ & $\hat{\varepsilon}_{{\rm out}}^{\ell}$ & $\hat{\varepsilon}^{\ell}$ & $\hat{\varepsilon}_{{\rm out}}^{\ell}$ & $\hat{\varepsilon}^{\ell}$ & $\hat{\varepsilon}_{{\rm out}}^{\ell}$ \\
\midrule
1 & 0.93 & 0.46 & 0.69 & 0.68 & 0.18 & 0.17 & 0.00 & 0.00 \\
2 & 0.81 & 0.13 & 0.17 & 0.28 & 0.07 & 0.19 & 0.00 & 0.01 \\
3 & 0.75 & 0.04 & 0.03 & 0.04 & 0.03 & 0.16 & 0.00 & 0.01 \\
4 & 0.72 & 0.04 & 0.03 & 0.00 & 0.03 & 0.12 & 0.00 & 0.01 \\
5& 0.72 & 0.02 & 0.01 & 0.00 & 0.03 & 0.08 & 0.00 & 0.02 \\
6 & 0.66 & 0.02 & 0.01 & 0.00 & 0.03 & 0.07 & 0.00 & 0.01 \\
7 & 0.64 & 0.02 & 0.01 & 0.00 & 0.03 & 0.05 & 0.01 & 0.02 \\
8 & 0.57 & 0.02 & 0.01 & 0.00 & 0.03 & 0.04 & 0.01 & 0.02 \\
9 & 0.51 & 0.02 & 0.01 & 0.00 & 0.04 & 0.03 & 0.01 & 0.02 \\
10 & 0.46 & 0.02 & 0.01 & 0.00 & 0.04 & 0.02 & 0.01 & 0.01 \\
11 & 0.40 & 0.02 & 0.01 & 0.00 & 0.04 & 0.02 & 0.01 & 0.02 \\
12 & 0.46 & 0.02 & 0.01 & 0.00 & 0.04 & 0.03 & 0.01 & 0.04 \\
13 & 0.55 & 0.02 & 0.00 & 0.00 & 0.03 & 0.02 & 0.01 & 0.03 \\
14 & 0.41 & 0.02 & 0.01 & 0.00 & 0.04 & 0.01 & 0.02 & 0.03 \\
15 & 0.67 & 0.02 & 0.00 & 0.00 & 0.03 & 0.01 & 0.02 & 0.08 \\
16 & 0.55 & 0.02 & 0.00 & 0.00 & 0.04 & 0.00 & 0.02 & 0.15 \\
17 & 0.42 & 0.02 & 0.00 & 0.00 & 0.04 & 0.00 & 0.03 & 0.09 \\
18 & 0.62 & 0.02 & 0.00 & 0.00 & 0.03 & 0.00 & 0.03 & 0.13 \\
19 & 0.17 & 0.01 & 0.00 & 0.00 & 0.06 & 0.00 & 0.20 & 0.05 \\
20 & 0.17 & 0.01 & 0.00 & 0.00 & 0.06 & 0.00 & 0.27 & 0.06 \\
21 & 0.16 & 0.01 & 0.00 & 0.00 & 0.06 & 0.00 & 0.33 & 0.20 \\
\bottomrule
\end{tabular}
\vspace{0.2cm}
\caption{Illustration of the evolution of the probability vector of classification for the different heads on ImageNET dataset with Convnext architecture and for three different budgets: each line represents an exit head, $\boldsymbol{\pi} $ is the prior distribution, $\hat{\boldsymbol{\varepsilon}} $ are the weights produced by the aggregation, and $\hat{\boldsymbol{\varepsilon}}_{{\rm out}} $ are the proportion of the data that was actually classified by each head.}
\label{epsilon_table}
\end{table*}
The results highlights that: i) when higher budget are allocated, the weights produced by the aggregation put higher weights on heads with lower risks; ii) heads with high risks are assigned smaller weights by the aggregation, see for instance the weights corresponding to the heads 13 and 18 that have a higher risk and so a lower value for the $\varepsilon^{\ell}$ threshold.

\begin{figure}

    \begin{subfigure}[b]{0.48\linewidth}
        \includegraphics[width=\linewidth]{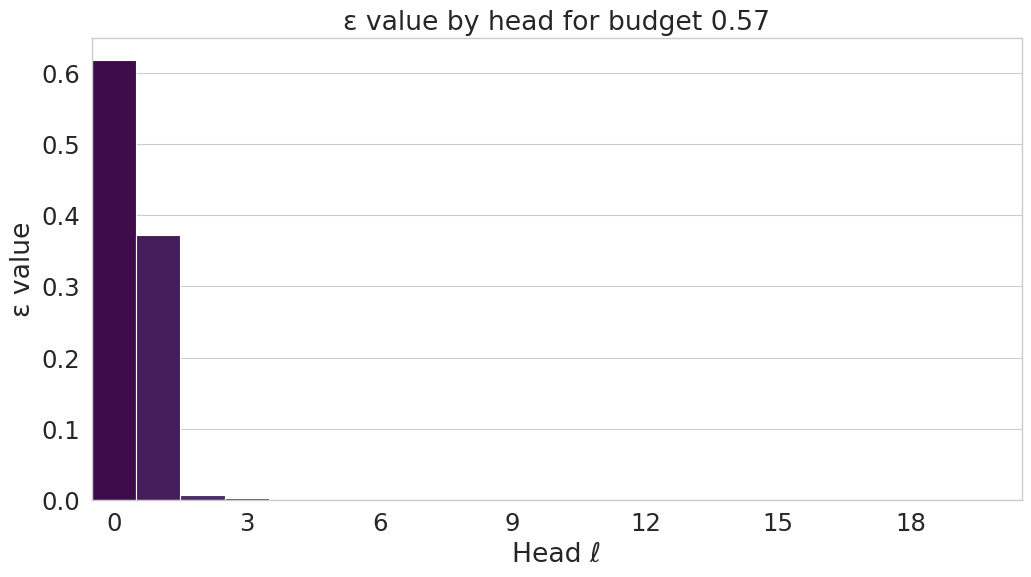}
    \end{subfigure}
    \hfill
        \begin{subfigure}[b]{0.48\linewidth}
        \includegraphics[width=\linewidth]{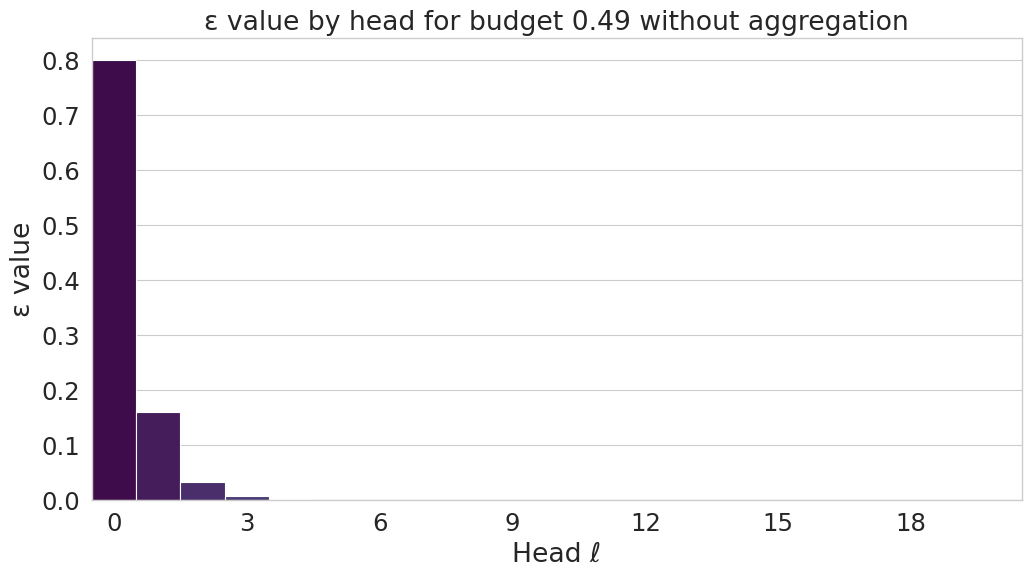}
    \end{subfigure}
    
    \begin{subfigure}[b]{0.48\linewidth}
        \includegraphics[width=\linewidth]{eps_budg_2.png}
    \end{subfigure}
    \hfill
    \begin{subfigure}[b]{0.48\linewidth}
        \includegraphics[width=\linewidth]{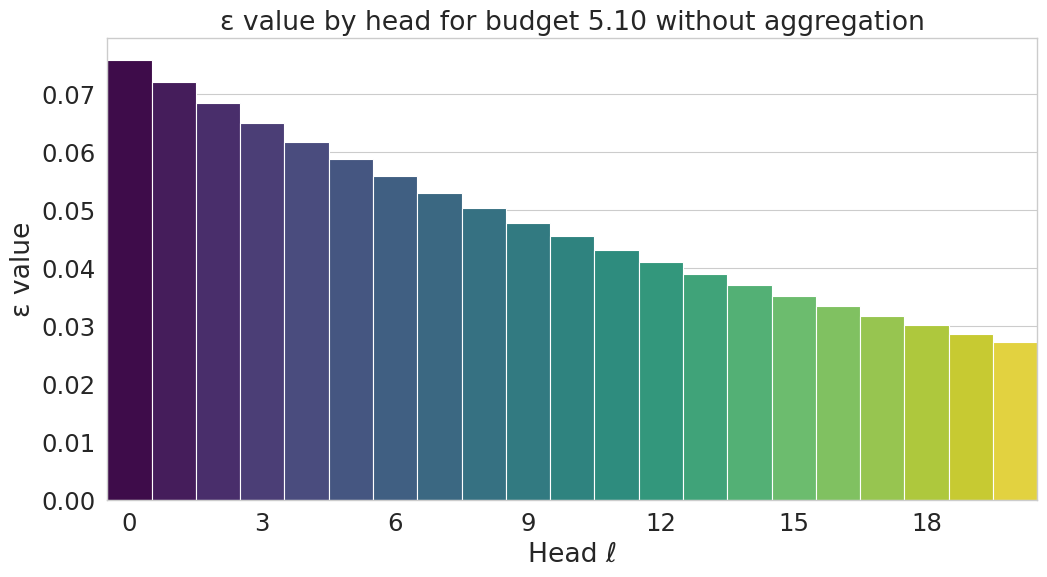}
    \end{subfigure}

    \begin{subfigure}[b]{0.48\linewidth}
        \includegraphics[width=\linewidth]{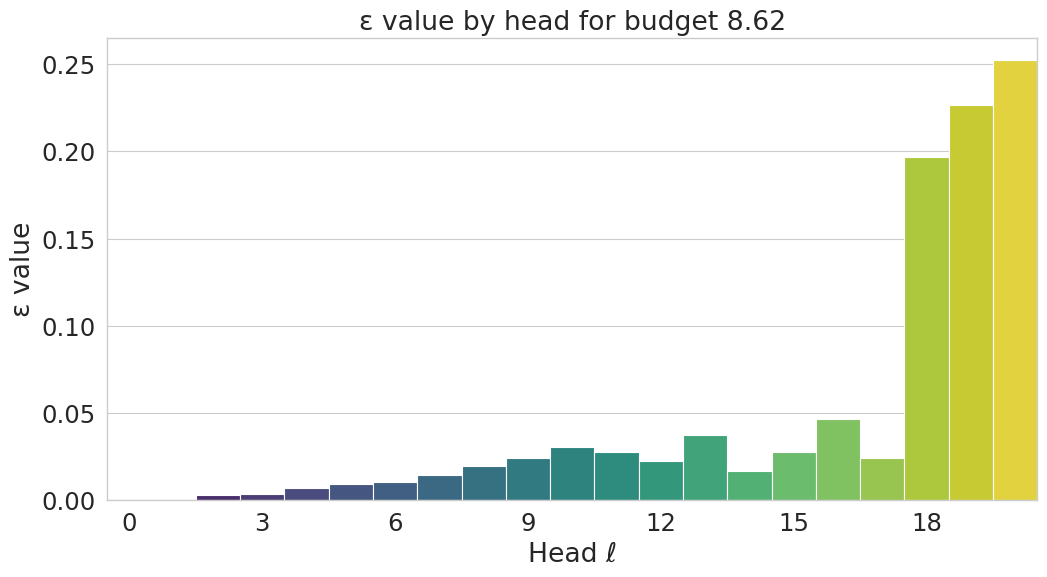}
    \end{subfigure}
    \hfill
    \begin{subfigure}[b]{0.48\linewidth}
        \includegraphics[width=\linewidth]{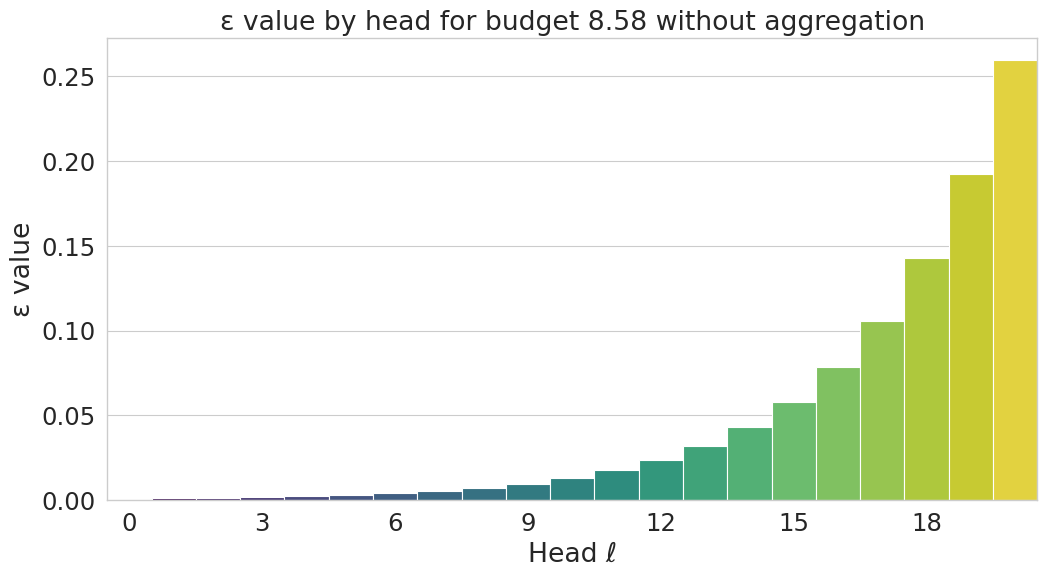}
    \end{subfigure}
        \centering  
    \caption{Value of the aggregation weights $\hat{\varepsilon}^{\ell}$ on three different budgets for the ConvNext model for EERO (left column) and with the MSDNet strategy (right column). Each bar represents an exit head $\ell$. The lines corresponds to different budgets ranging from small (top), to medium (centre) and high~(top).}
\label{fig:epsilon_per_budget}
\end{figure}

This is also visible in Figure~\ref{fig:epsilon_per_budget} where we show the $\hat{\boldsymbol{\varepsilon}}$ values for our EERO method together with the MSDNet like strategy.
\begin{figure}[t]

    \subfloat[Accuracy \emph{w.r.t.} the real budget based on MSDNet architecture -- comparison of EERO methodology to the original MSDNet algorithm.\label{fig:MSDNet_acc}]
    {\includegraphics[width=.45\linewidth]{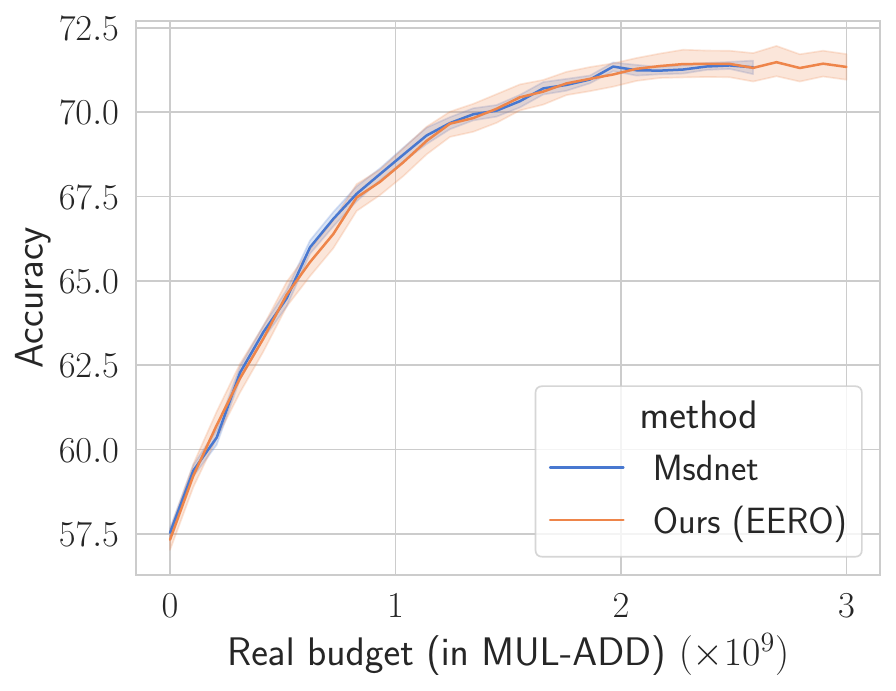}}
    \hfill
    \subfloat[Measured budget on MSDNet architecture -- comparison of EERO methodology to the original MSDNet algorithm.\label{fig:MDSNet_budget}]
    {\includegraphics[width=.45\linewidth]{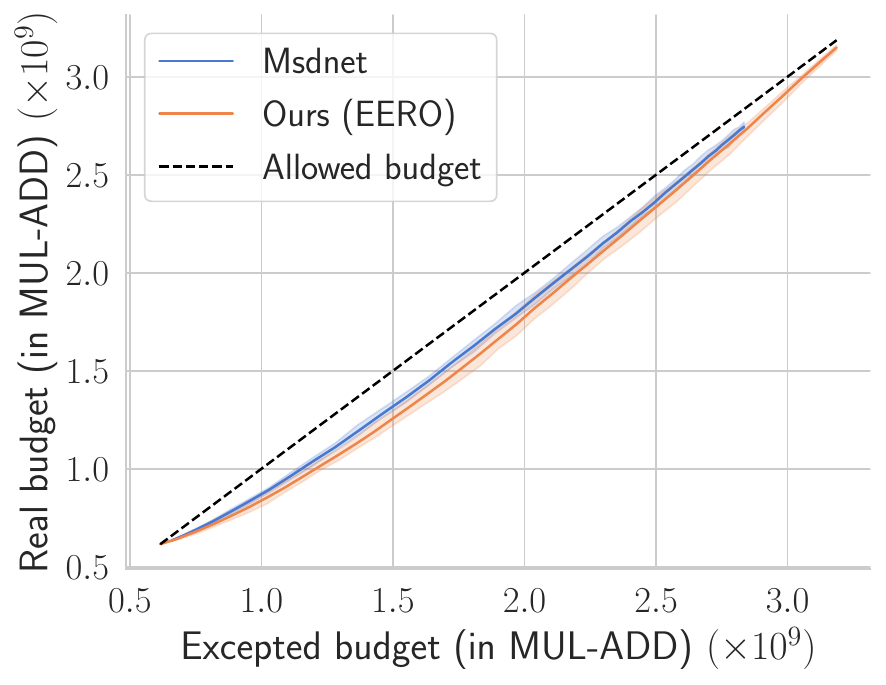}}
    
\caption{Subfigures (a) and (b) compare EERO with MSDNet's original method, focusing on accuracy and budget metrics. Error bars were made using bootstrap on data.}
\end{figure}

\begin{figure}
    \centering
    \begin{subfigure}[b]{0.75\linewidth}
        \includegraphics[width=\linewidth]{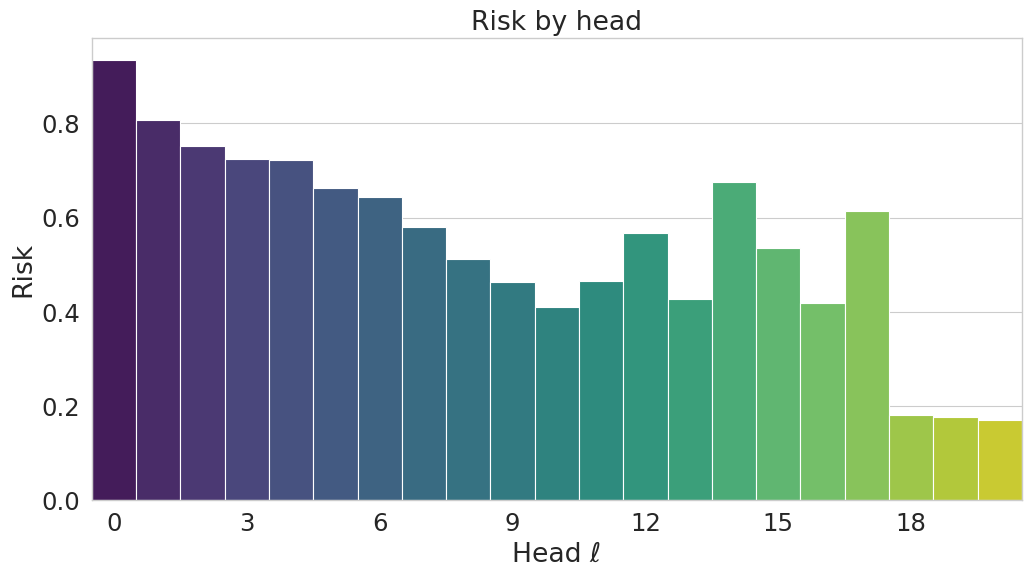}
    \end{subfigure}
    \caption{Risk $\hat{R}^{\ell}$ of each head on Convnext model}
\label{fig:risk}
\end{figure}

First line shows a budget that does not allow usage of the computationally costly heads and so most data will be output on the two first heads. Second line corresponds a medium budget, where we can see that our EERO approach and the MSDNet strategy show different behaviors. Whereas the latter assumes that a sample that \textit{reaches} a head classifier has the same constant probability to exit independently from its layer position, resulting in a monotonic behavior, our approach exploits the information of the empirical risks $\hat{R}^{\ell}$ (see Equation~\eqref{eq:pbExpWeigh}) to modulate the importance of the heads. As we saw in Figure~\ref{fig:epsilon_per_budget} and~\ref{fig:risk}, a high risk is correlated with a low epsilon. Third line shows a permissive budget where most data will be output at the last heads as they have more accuracy.

\section{EERO with only 1 auxiliary head}
\label{app:TwoHeads}

\begin{figure}
    \centering
    \resizebox{0.45\textwidth}{!}{ \includegraphics{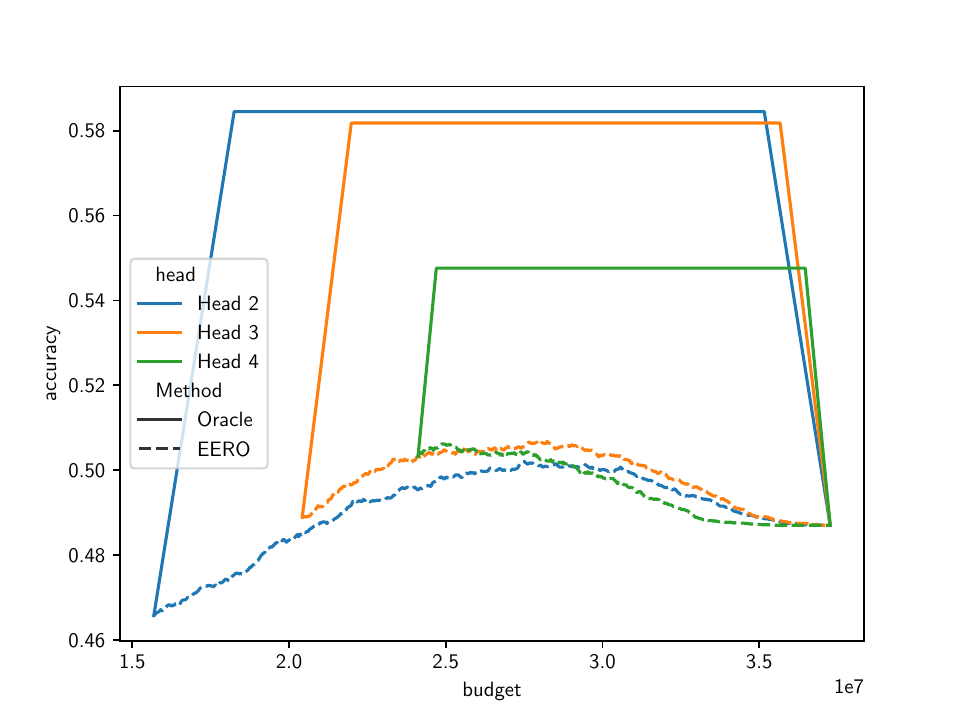}}
    \caption{Accuracy versus computation budget for our multi-headed ResNet-18 (given in Figure~\ref{fig:resnet}). In this illustration, we consider only a two heads procedure by selecting a specific auxiliary head -- 2, 3, or 4 -- among the initial 7. Each curve shows how the accuracy evolves according to the probability of use of the auxiliary head. For each color, we display our algorithm as well as the oracle.}
\label{fig:accuracy_budget}
\end{figure}

In the case of one auxiliary head, computations of the rate of classifying $\hat{\varepsilon}^1$ (of this single head) can be simplified. 
In particular, we can express it analytically without the need for aggregation with exponential weights.
Let $B>0$ be the amount of \emph{GFlops} we were allocated for the task of labeling $T$ data points.
Assume that the classifiers $\hat{h}_{\hat{\varepsilon}^1}^1$ and $\hat{g}^2$ burn off respectively $\hat{B}^1$ and $\hat{B}^2$ \emph{GFlops} at each call with $\hat{B}^1 < \hat{B}^2 $. (We assume that
$B \geq T \hat{B}^1$ so that we are guaranteed to label all $T$ points). 
\begin{proposition}
\label{prop:rejectionOneHead}
    Let $\hat{\varepsilon}^1 = \frac{\frac{B}{T}-\hat{B}^2}{\hat{B}^1 - \hat{B}^2}$. Then the total amount of \emph{GFlops} used to label $T$ instances is not larger than $B$.
\end{proposition}
This result highlights that our strategy succeeds to comply with the constraint of budget we imposed. The proof of the result lies in the fact that we use the early exit on a proportion $\hat{\varepsilon}^1$ of the data. Then we consume $T \hat{\varepsilon}^1 \hat{B}^1$ \emph{GFlops} (up to a $1/\sqrt{N}$ additive term -- see the comment after Proposition~\ref{propo:EmpiricalRejectionRate} for a correction). The rest of the data is treated by the classifier $\hat{g}^2$ and then uses $T (1-\hat{\varepsilon}^1) \hat{B}^2$ \emph{GFlops}. In total, we then get
\begin{equation}
\label{eq:burnedbugetonehead}
    \begin{split}
        T (\hat{\varepsilon}^1 \hat{B}^1 +  (1-\hat{\varepsilon}^1) \hat{B}^2)
         = B \enspace.
     \end{split}
\end{equation}
\begin{remark}
\label{rq:overthinking}
In this section, 
    we enforce the rejection rate to exactly burn off the whole budget $B$ -- see Equation~\eqref{eq:burnedbugetonehead}. We make this choice to explain some broader effects that illustrate the importance of using auxiliary heads (see below and Figure~\ref{fig:accuracy_budget}).
    On the other hand, we notify that it is extremely simple to modify the algorithm so that it consumes less than or equal to the total budget, that is, $(\hat{\varepsilon}^1 T) \hat{B}^1 + ((1-\hat{\varepsilon}^1)T) \hat{B}^2 \leq  B$. In this case, the accuracy curve would always be increasing \emph{w.r.t.} the budget.
\end{remark}


We implement EERO using a ResNet~\citep{he2016deep} model on the CIFAR-100~\citep{krizhevsky2009learning} dataset, with auxiliary heads placed at seven different positions for early exit testing (see Figure~\ref{fig:resnet}). The model training involve a sum of cross-entropy losses from these exits. Detailed hyperparameters of this experiment are provided in Appendix~\ref{app:details} in Tables~\ref{table:basic-setup} and~\ref{table:training-parameters}.

We plot the accuracy for the different exits in Figure~\ref{fig:accuracy_budget} for different budgets. 
First, we can see a general trend where our procedure gradually improves the accuracy as the budget increases since images wrongly classified by the auxiliary head and correctly classified by the final layer are forwarded from the Early Exit to the latter. Then there is an accuracy decrease since augmenting the budget only affects images which are wrongly classified by both Early Exit and final model exit. Secondly, we notice that the oracle quickly outperforms the accuracy of the main model for only a fraction of the total budget required to process all images with the full model.
This fact shows that Early Exits can have valid predictions, whereas the last model output is wrong. This phenomenon is known as over-thinking and motivates the potential of Early Exit methods, as a way of better understanding the flaws of deep learning models. Here, this over-thinking situation seems important, as some of the Early Exits have a better accuracy than the last model output, \emph{c.f.,} Remark~\ref{rq:overthinking}. A second reason explaining the oracle improvements is its capacity to save computation budget by choosing an Early Exit when both early and last exit predictions are wrong.

This advantage, only achievable in our specific context of Budgeted Batch Classification, aligns well with our primary objective of reducing computational power for our algorithms. Furthermore, it indirectly enhances the overall accuracy of our model. 

\section{Proofs}
\label{App:proofs}
In this section, we provide the proofs of our main results.
\begin{proo}[Proposition~\ref{prop:BayesRejectRule}]
Let us write 
\begin{equation*}
    \mathcal{R}(h) = \mathbb{P}\left( h(\bX) \neq Y \ , \ h(\bX) \neq \mathfrak{R}  \right)\enspace,
\end{equation*}
for short. We then need to solve the problem 
\begin{equation*}
 h^*_{\varepsilon} = \argmin{h} \left\{ \mathcal{R}(h) : \mathbb{P}(h(\bX) = \mathfrak{R} ) = 1- \varepsilon \right\} ,
\end{equation*}
where the minimum is taken over all measurable functions.
Considering the Lagrangian of the above problem, we solve
\begin{equation*}
 \min_{h} \max_{\lambda\in \mathbb{R}^+} \underbrace{\left\{ \mathcal{R}(h) +\lambda \left(  \mathbb{P}(h(\bX) = \mathfrak{R} ) - (1 - \varepsilon)  \right) \right\} }_{:=\mathcal{L}(h,\lambda)} \enspace,
\end{equation*}
where $\lambda \in \mathbb{R}$ is the dual variable.
Observe that by weak duality we have
$$\min_{h} \max_{\lambda\in \mathbb{R}^+}  \mathcal{L}(h,\lambda) \geq \max_{\lambda\in \mathbb{R}^+} \min_{h} \mathcal{L}(h,\lambda)\enspace,$$ we then consider first the minimization problem over $h$ of $\mathcal{R}$.
We have 
\begin{multline*}
    \mathcal{R}(h) = \mathbb{P}\left( h(\bX) \neq Y \ , \ h(\bX) \neq \mathfrak{R}  \right) 
    =  \mathbb{E}\left[ \sum_{k\in[K]} \one{Y=k} \one{ h(\bX) \neq k}  \one{h(\bX) \neq \mathfrak{R} } \right] \\
     = \mathbb{E}\left[\sum_{k\in[K]} p_{k}(\bX) \one{ h(\bX) \neq k} \one{h(\bX) \neq \mathfrak{R} }\right]
     = \mathbb{P}\left( h(\bX) \neq \mathfrak{R}  \right)  - \mathbb{E}\left[\sum_{k\in[K]} p_{k}(\bX) \one{ h(\bX) = k} \one{h(\bX) \neq \mathfrak{R} }\right]
\end{multline*}
Moreover,
\begin{equation*}
    \mathbb{P}(h(\bX) \neq \mathfrak{R} ) =  \mathbb{E}\left[ \sum_{k\in [K]}  \one{h(\bX) = k } \one{h(\bX) \neq \mathfrak{R} } \right] . 
\end{equation*}
Therefore, we can write
\begin{eqnarray}
\label{eqproof:biglagrangian}
    \mathcal{L}(h,\lambda) & = &  \lambda - \lambda (1-\varepsilon)  + (1-\lambda)  \mathbb{P}(h(\bX) \neq \mathfrak{R} ) 
    - \mathbb{E}\left[\sum_{k\in[K]} p_{k}(\bX) \one{ h(\bX) = k} \one{h(\bX) \neq \mathfrak{R} }\right] \\ 
   &  = & \lambda  \varepsilon  - 
    \mathbb{E}\left[  \sum_{k\in [K]} \left (p_k(\bX) - (1-\lambda) \right) \one{h(\bX) = k } \one{h(\bX) \neq \mathfrak{R} } \right]  \enspace.
\end{eqnarray}
We need to maximize the expectation \emph{w.r.t.} $h$ that first leads to the optimum 
$h_{\lambda}^*$ is such that 
\begin{equation}
    h_{\lambda}^*(\bx) \neq  \mathfrak{R}  \iff  \sum_{k\in [K]} \left (p_k(\bx) - (1-\lambda) \right) \one{h_{\lambda}^*(\bx) = k } >0 \enspace,
\end{equation}
for all $\bx \in \mathcal{X}$. Moreover we have that on the event $\left\{ h_{\lambda}^*(\bx) \neq  \mathfrak{R}\right\} $, the mapping $h$ that maximizes $h\mapsto \sum_{k\in [K]} \left (p_k(\bx) - (1-\lambda) \right) \one{h(\bx) = k } $ is simply $h_{\lambda}^*(\bx) = \argmax{k\in[K]}{p_k}(\bx)$.
At this level of the proof, we have shown that the problem $\min_h \mathcal{L}(h,\lambda) $ leads to the rule
\begin{equation}
\label{eqproof:optimalrulepartial}
    h_{\lambda}^* (\bx)   =
    \left\{
    \begin{array}{ll}
        \argmax{k\in [K]} p_k(\bx)         
         & \text{if} \quad \max_{k\in[K]} p_k(\bx) \geq 1-  \lambda
         \\
        \mathfrak{R} & \text{otherwise}\enspace.
    \end{array} 
    \right.
\end{equation}
Now, we deal with the maximization of $\mathcal{L}(h_{\lambda}^*,\lambda)$ \emph{w.r.t.} $\lambda$.
Substituting the above value of $h=h_{\lambda}^* $ in~\eqref{eqproof:biglagrangian}, we can show that
\begin{equation*}
    \mathcal{L}(h_{\lambda}^*,\lambda) =
    \lambda \varepsilon -
    \mathbb{E}\left[  \left( \max_{k\in[K]}   \left\{ p_k(\bX) - (1-\lambda) \right\} \right)_+ \right] \enspace,
\end{equation*}
where for all $a\in \mathbb{R}$, we write $(a)_+ = \max\{a, 0\}$. The above function is then concave in $\lambda$, therefore we can write the first order optimality condition as $0 \; \in \partial \mathcal{L}(h_{\lambda^*}^*,\lambda^*)$. Observe that because we assumed that the r.v. $p_k(X)$ has no atom for all $\in[K]$, we have that $\mathbb{P} \left( \exists \; j\in[K]: p_k(X) = p_j(X) \right) = 0$ and then the subgradient reduces to the gradient. As a consequence, we obtain the following condition on $\lambda^*$
\begin{equation*}
    \mathbb{P} \left( \exists k\in [K] :
     p_k(\bX) > \max \left\{ \max_{j\in[K]} \left\{p_j(\bX)\right\} ; 1-\lambda^* \right\}
    \right) 
    = \varepsilon \enspace .
\end{equation*}
Notice that this last condition can rewrite as
\begin{equation*}
    \varepsilon  = \mathbb{P} \left( s(\bX)
     >  1-\lambda^* 
    \right) =  \mathbb{P} \left(  h_{\lambda^*}^* (\bX) \neq \mathfrak{R} 
    \right)  \enspace ,
\end{equation*}
where $ s(\bX) =   \max_{k\in[K]} p_k(\bX)$, which guarantee that the optimal rule has indeed the correct rejection rate. Moreover, from the above relation, we can exhibit the value of $\lambda$. Indeed, using the continuity condition on the CDF $F_s$ of $s(\bX)$, we have
\begin{equation*}
    \mathbb{P} \left( s(\bX)
     >  1-\lambda^*\right) = \varepsilon 
     \quad  \iff   \quad 
     1-F_{s}\left( 1-\lambda^* \right) = \varepsilon
     \quad 
      \iff
     \quad 
      \lambda^*= 1 -  F_{s}^{-1}\left( 1- \varepsilon  \right), 
\end{equation*}
where $F_{s}^{-1}$ is the generalized inverse of $F_{s}$. We conclude the proof substituting this value into the expression of the optimal rule given by~\eqref{eqproof:optimalrulepartial}.

\begin{remark}
    If we replace the equality by an inequality, everything is the same expect the part of $\lambda^*$. We need to consider the case where $\lambda^* = 0$ separately, and in this case, we would get $\mathbb{P} \left(  h_{\lambda^*}^* (\bX) \neq \mathfrak{R} 
    \right) \geq \varepsilon$.
\end{remark}

\end{proo}

\begin{proo}[Proposition~\ref{propo:EmpiricalRejectionRate}]
First, observe that conditionally to the training dataset $\mathcal{D}_n$ and due to the continuity of the CDF of $\hat{s}^{\ell}(\bX)$, the random variable $F_{\hat{s}^{\ell}}(\hat{s}^{\ell}(\bX)) $ is uniformly distributed. Therefore, for any $u\in [0,1]$, we have $\mathbb{P}\left( F_{\hat{s}^{\ell}}(\hat{s}^{\ell}(\bX)) \leq u \right) = u$. We then can write 
\begin{multline*}
\left| \mathbb{P}\left( \hat{h}_{\varepsilon^{\ell}}^{\ell}(\bX) =  \mathfrak{R} \right) - (1-\varepsilon^{\ell}) \right| 
 = 
\left| \mathbb{E}  \left[ \one{\hat{F}_{\hat{s}^{\ell}}(\hat{s}^{\ell}(\bX))  \leq 1-\varepsilon^{\ell}} -
\one{F_{\hat{s}^{\ell}}(\hat{s}^{\ell}(\bX))  \leq 1-\varepsilon^{\ell}} 
\right]  \right| 
\\ \leq 
\left| \mathbb{E}  \left[ \one{ \left|F_{\hat{s}^{\ell}}(\hat{s}^{\ell}(\bX)) - (  1-\varepsilon^{\ell}  ) \right| \leq  \left|\hat{F}_{\hat{s}^{\ell}}(\hat{s}^{\ell}(\bX)) - F_{\hat{s}^{\ell}}(\hat{s}^{\ell}(\bX))   \right|  
 } 
\right]  \right| 
\\ \leq 
\left| \mathbb{E}  \left[ \one{ \left|F_{\hat{s}^{\ell}}(\hat{s}^{\ell}(\bX)) - (  1-\varepsilon^{\ell}  ) \right| \leq  \left\Vert \hat{F}_{\hat{s}^{\ell}} - F_{\hat{s}^{\ell}}   \right\Vert_{\infty}  
 } 
\right]  \right| 
 \leq 
2 \mathbb{E}   \left\Vert \hat{F}_{\hat{s}^{\ell}} - F_{\hat{s}^{\ell}}   \right\Vert_{\infty} \enspace, 
\end{multline*}
where we used again in the last line the fact that $F_{\hat{s}^{\ell}}(\hat{s}^{\ell}(\bX)) $ is uniformly distributed. Moreover, we wrote $\Vert  \hat{F}_{\hat{s}^{\ell}} - F_{\hat{s}^{\ell}} \Vert_{\infty} =  \sup_{t \in \mathbb{R}}\vert 
\hat{F}_{\hat{s}^{\ell}}(t) - F_{\hat{s}^{\ell}}(t) \vert $. We conclude the proof using the Dvoretzky–Kiefer–Wolfowitz inequality,
that states that 
\begin{equation*}
    \mathbb{E}   \left\Vert \hat{F}_{\hat{s}^{\ell}} - F_{\hat{s}^{\ell}}   \right\Vert_{\infty}
    \leq \sqrt{\frac{\pi}{2N}}\enspace.
\end{equation*}

\end{proo}

\begin{proo}[Proposition~\ref{prop:rejectionratedefinition}]
The Lagrangian of the minimization problem~\eqref{eq:pbExpWeigh}-\eqref{eq:pbExpWeighConstraint} is given by 
\begin{equation*}
    \mathcal{L}(\boldsymbol{\varepsilon},\lambda,\mu) = \sum_{j=1}^M  \varepsilon^j \hat{R}^{j} + \beta \sum_{j=1}^M \varepsilon^j \log \left( \frac{\varepsilon^j}{\pi^j} \right) 
    + \lambda \left( 
    \sum_{j=1}^M  \varepsilon^j -1  
    \right) 
    + \mu \left( 
    \sum_{j=1}^M  \varepsilon^j \hat{B}^{j}  - \bar{B}
    \right)\enspace,
\end{equation*}
with $(\lambda,\mu) \in \mathbb{R} \times \mathbb{R}^+$. Considering the KKT condition of the problem we get 
for all $j \in [M]$
\begin{equation*}
    \hat{R}^{j}  + \beta \left( \log \left( \frac{ \hat{\varepsilon}^j}{\pi^j} \right) +1 \right) + \hat{\lambda} + \hat{\mu} \hat{B}^{j} = 0,
\end{equation*}
that leads in turns to
\begin{equation}
\label{eqproof:hatepsilon}
    \hat{\varepsilon}^j_{\hat{\lambda} , \hat{\mu} } = \pi^j \exp \left( -\frac{\hat{R}^{j}  +   \hat{\lambda} + \hat{\mu} \hat{B}^{j}}{\beta} - 1 \right) \enspace.
\end{equation}
Plug-in these values into the equality constraint leads to
\begin{eqnarray*}
    \sum_{j=1}^M  \hat{\varepsilon}_{\hat{\lambda} , \hat{\mu} }^j =1 
    &  \iff &
    \sum_{j=1}^M \pi^j \exp \left( -\frac{\hat{R}^{j} + \mu \hat{B}^{j}}{\beta} - 1 \right) = \exp\left(  \frac{\hat{\lambda}}{\beta} \right)
    \\
    & \iff &
    \hat{\lambda} =
    \beta \log \left( \sum_{j=1}^M \pi^j \exp \left( -\frac{\hat{R}^{j} + \mu \hat{B}^{j}}{\beta} - 1 \right) \right)\enspace.
\end{eqnarray*}
Substituting back this value into~\eqref{eqproof:hatepsilon}, we get
\begin{equation*}
    \hat{\varepsilon}^j_{ \hat{\mu} } = \frac{\pi^j \exp \left( -\frac{\hat{R}^{j}  +    \hat{\mu} \hat{B}^{j}}{\beta} \right)}{\sum_{k=1}^M \pi^k \exp \left( -\frac{\hat{R}^{k}  +    \hat{\mu} \hat{B}^{k}}{\beta}  \right) }\enspace.
\end{equation*}
According to the parameter $\hat{\mu}$, we need to consider the constraints $\hat{\mu} \geq 0$ and $\sum_{j=1}^M  \hat{\varepsilon}_{\hat{\mu}}^j \hat{B}^{j} \leq \bar{B}$ together with the complementary condition $\hat{\mu} \left( \sum_{j=1}^M  \hat{\varepsilon}_{\hat{\mu}}^j (\hat{B}^{j} - \bar{B} ) \right)  = 0$. Therefore, when $\hat{\mu} \neq 0$, this parameter should be taken such that
\begin{equation*}
    \sum_{j=1}^M  \hat{\varepsilon}_{\hat{\mu}}^j (\hat{B}^{j} -\bar{B}) = 0
     \quad  \iff  \quad 
     \sum_{j=1}^M \pi^j (\hat{B}^{j} - \bar{B}) \exp \left( -\frac{\hat{R}^{j}  +    \hat{\mu}  \hat{B}^{j}}{\beta} \right) 
     = 0 \enspace.
\end{equation*}
Otherwise $\hat{\mu} = 0$ and in this case, $ \hat{\varepsilon}^j_{ \hat{\mu} }$ becomes
\begin{equation*}
    \hat{\varepsilon}^j_{ \hat{\mu} } = \frac{\pi^j \exp \left( - \hat{R}^{j}  / \beta \right)}{\sum_{k=1}^M \pi^k \exp \left( -\hat{R}^{k}   / \beta \right) }\enspace.
\end{equation*}

\end{proo}

\begin{proo}[Theorem~\ref{thm:OracleInequality}]
Let us first give a formal definition of the resulting classifier $\hat{g}^{\text{EERO} }$ from our procedure. Introduce for all $\bx \in  \mathcal{X}$, the index $\bar{\ell}(\bx) = \min \left\{ \ell \in [M]: \ \hat{F}_{\hat{s}^{\ell}}(\hat{s}^{\ell}(\bx))  \geq 1 -\hat{\varepsilon}^{\ell}  \right\}$ as the first moment where the we do not reject. Then, the classification by EERO is exactly provided by the head that corresponds to $\bar{\ell}$, that is
\begin{eqnarray}
    \label{proofdef:EERO}
    \hat{g}^{\text{EERO} }(\bx) = \hat{g}^{\bar{\ell}(\bx)} (\bx)\enspace.
\end{eqnarray}
The goal now is here to demonstrate that the choice of $\hat{\boldsymbol{\varepsilon}}$ provides a good compromise in terms of risk and computations budget. Recall that for all $\ell\in [M]$, we have 
    \begin{equation*}
    \hat{\varepsilon}^{\ell} = \frac{\pi^{\ell} \exp{\left\{ - \frac{\hat{R}^{\ell} + \hat{\mu} \hat{B}^{\ell}}{\beta} \right\} }}{ \sum_{j=1}^M \pi^j \exp{\left\{ - \frac{\hat{R}^j + \hat{\mu} \hat{B}^j}{\beta} \right\}} }  = c \cdot \pi^{\ell} \exp{\left\{ - \frac{\hat{R}^{\ell} + \hat{\mu} \hat{B}^{\ell}}{\beta} \right\} } \enspace,
\end{equation*}
where $c =\sum_{j=1}^M \pi^j \exp{\left\{ - \frac{\hat{R}^j + \hat{\mu} \hat{B}^j}{\beta} \right\}}  $ is the normalization constant. Then 
$$
\log\left( \frac{\hat{\varepsilon}^{\ell} }{ \pi^{\ell}} \right)= \log(c) - \frac{\hat{R}^{\ell} + \hat{\mu} \hat{B}^{\ell}}{\beta} \enspace.
$$
Rearranging terms we get 
$$
\hat{R}^{\ell} 
=   \beta \log(c) - \beta \log\left( \frac{\hat{\varepsilon}^{\ell} }{ \pi^{\ell}} \right) - \hat{\mu} \hat{B}^{\ell}\enspace.
$$
Notice that the same relation holds true for the optimal head $\ell^*$ -- the feasible head that minimizes the risk. Therefore, subtracting the two equality gives
$$
\hat{R}^{\ell} = \hat{R}^{\ell^*} +  \beta \left( \log\left( \frac{\hat{\varepsilon}^{\ell^*} }{ \pi^{\ell^*}} \right) -   \log\left( \frac{\hat{\varepsilon}^{\ell} }{ \pi^{\ell}} \right)   \right) +  \hat{\mu} \left( \hat{B} ^{\ell^*}  - \hat{B} ^{\ell} \right) \enspace.
$$
Multiplying by $\hat{\varepsilon}^{\ell}$ and summing out over $\ell$ we get
\begin{equation}
    \label{proofeq:IntermediateEquality}
    \frac{1}{n_2} \sum_{(\bX,Y) \in \mathcal{D}_{n_2} }
\one{\hat{g}^{\text{EERO} } (\bX)  \neq  Y } =
\sum_{\ell = 1}^{M}\hat{\varepsilon}^{\ell} 
\hat{R}^{\ell} 
= 
\hat{R}^{\ell^*} +  \beta \left( \log\left( \frac{\hat{\varepsilon}^{\ell^*} }{ \pi^{\ell^*}} \right) -   \sum_{\ell = 1}^{M}\hat{\varepsilon}^{\ell} \log\left( \frac{\hat{\varepsilon}^{\ell} }{ \pi^{\ell}} \right)   \right) +  \hat{\mu} \left( \hat{B} ^{\ell^*}  -\sum_{\ell = 1}^{M}\hat{\varepsilon}^{\ell}   \hat{B} ^{\ell} \right) \enspace.
\end{equation}
Let us first consider the term $ \hat{\mu} \left( \hat{B} ^{\ell^*}  -\sum_{\ell = 1}^{M}\hat{\varepsilon}^{\ell}   \hat{B} ^{\ell} \right)$. Recalling the previously stated complementary condition $\hat{\mu} \left( \sum_{\ell =1}^M  \hat{\varepsilon}^{\ell} (\hat{B}^{\ell} - \bar{B} ) \right)  = 0$, we either have
$\hat{\mu} = 0$ or $\sum_{\ell =1}^M  \hat{\varepsilon}^{\ell}  \hat{B}^{\ell} = \bar{B}$. In the latter case, the term 
$$ \hat{B} ^{\ell^*}  -\sum_{\ell = 1}^{M}\hat{\varepsilon}^{\ell}   \hat{B} ^{\ell} =\hat{B} ^{\ell^*} - \bar{B} \leq 0 \enspace, 
$$ 
since $\hat{B} ^{\ell^*} \leq \bar{B} $ by the fact that it corresponds to the budget of the $\ell$-th head with a weight $\hat{\varepsilon}^* = 1$ (all other weights are $0$). Therefore, in both cases $ \hat{\mu} \left( \hat{B} ^{\ell^*}  -\sum_{\ell = 1}^{M}\hat{\varepsilon}^{\ell}   \hat{B} ^{\ell} \right) \leq 0$. 
On the other, we have by the properties of the KL divergence,
$$   
\sum_{\ell = 1}^{M}\hat{\varepsilon}^{\ell} \log\left( \frac{\hat{\varepsilon}^{\ell} }{ \pi^{\ell}} \right)   \leq 0 \enspace,
$$
and then Equation~\eqref{proofeq:IntermediateEquality} becomes
$$
\frac{1}{n_2} \sum_{(\bX,Y) \in \mathcal{D}_{n_2} }
\one{\hat{g}^{\text{EERO} } (\bX)  \neq  Y } \leq 
\hat{R}^{\ell^*} +   \beta     \log (  1/\pi^{\ell^*} ) \enspace,
$$
since $\log ( \hat{\varepsilon}^{\ell^*} ) = 0$ and this ends the proof by taking the expectation from both side.


\end{proo}

\end{document}